\documentclass{article} %
\usepackage{iclr2027_conference,times}

\usepackage{amsmath,amsfonts,bm}

\def\1{\bm{1}}

\DeclareMathAlphabet{\mathsfit}{\encodingdefault}{\sfdefault}{m}{sl}
\SetMathAlphabet{\mathsfit}{bold}{\encodingdefault}{\sfdefault}{bx}{n}

\newcommand{\R}{\mathbb{R}}

\newcommand{\softmax}{\mathrm{softmax}}

\usepackage{latexsym}
\usepackage[T1]{fontenc}
\usepackage[utf8]{inputenc}
\usepackage{microtype}
\usepackage{inconsolata}
\usepackage{graphicx}
\usepackage{subcaption}
\usepackage{booktabs}
\usepackage{xcolor}
\usepackage{colortbl}
\usepackage{amsmath}
\usepackage{amssymb}
\usepackage{multirow}
\usepackage{enumitem}
\newcommand{\rope}{\text{RoPE}}
\newcommand{\hrope}{\overline{\rope}}
\usepackage{hyperref}
\usepackage{url}

\title{GQLA: Group-Query Latent Attention \\for Hardware-Adaptive Decoding}

\author{%
\textbf{Fanxu Meng}\thanks{Equal contribution.}\\
Dots Studio, Xiaohongshu Inc.\\
\href{mailto:mengfanxu@xiaohongshu.com}{\texttt{mengfanxu@xiaohongshu.com}}
\And
\textbf{Zhixin Pan}\footnotemark[1]\\
Dots Studio, Xiaohongshu Inc.\\
Beijing Institute of Technology\\
\href{mailto:1120241050@bit.edu.cn}{\texttt{1120241050@bit.edu.cn}}
}
\hypersetup{pdfauthor={Fanxu Meng, Zhixin Pan}}

\iclrfinalcopy %
\makeatletter
\renewcommand{\@maketitle}{\vbox{\hsize\textwidth\centering
  \def\@makefnmark{\hbox{\@textsuperscript{\normalfont\@thefnmark}}}%
  {\LARGE\scshape\@title\par}
  \def\And{\end{tabular}\hspace{2em}%
    \begin{tabular}[t]{c}\rule{\z@}{24pt}\ignorespaces}%
  \begin{tabular}[t]{c}\rule{\z@}{24pt}\@author\end{tabular}\par
  \vskip 0.6em
  {\normalfont\small\url{https://github.com/MuLabPKU/TransArch}\par}
  \vskip 0.3in minus 0.1in
}}
\makeatother

\begin{document}

\maketitle
\lhead{} %

\begin{abstract}
Decoding efficiency depends on KV-cache traffic, computation, and parallelism.
Multi-head Latent Attention (MLA) offers absorbed and head-expanded forms, but its head-specific
projections limit grouped KV sharing. We introduce \textbf{Group-Query Latent
Attention (GQLA)}, whose group-shared up-projections preserve a joint latent
and enable a shardable expanded cache. The same weights support equivalent
MQA-absorb and GQA execution using existing kernels. Roofline analysis relates
path preference to GPU compute-to-bandwidth ratio and tensor parallelism (TP):
absorption trades traffic for computation and replicates the latent across ranks;
GQA shards KV groups. Pretraining with eight query heads per group yields
terminal training losses close to MLA's across the tested model scales at matched tokens.
\textbf{TransGQLA} converts GQA and MLA checkpoints using calibration data.
Without gradient updates, converted DeepSeek-V3.1-Base achieves $77.79\%$
eight-task accuracy versus the source's $77.84\%$, while WikiText-2 perplexity
rises from $3.21$ to $6.05$. On eight H20-3e GPUs, its GQA path improves output
throughput over MQA-absorb by $22.42$--$34.97\%$ at matched resident batches
across 8K--64K inputs. Its smaller per-rank cache permits larger batches,
giving $50.07\%$ in a separate capacity-tuned comparison. These results support
multiple execution paths as an architectural means of adapting decoding to
hardware and parallelism.

\end{abstract}

\begin{figure*}[bh]
  \centering
  \includegraphics[width=0.99\linewidth]{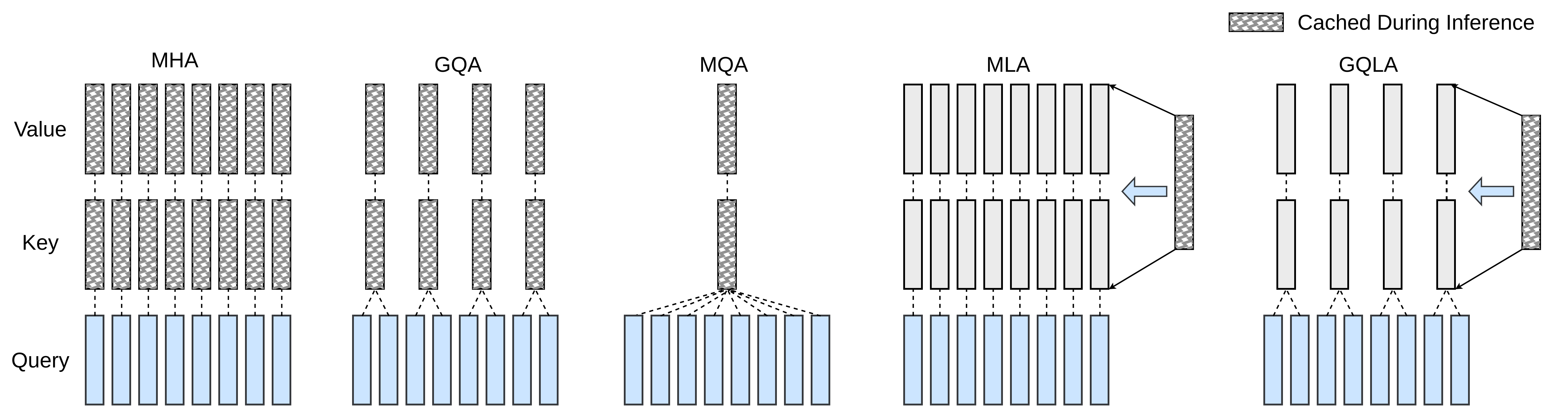}
  \caption{Attention representations and their cache structure.
GQLA adds group-shared up-projections to a joint latent, exposing both
compact MQA-absorb and expanded GQA execution over the same weights
(\S\ref{sec:gqla}).}
  \label{fig:mha_gqa_mqa_mla_gqla}
  \end{figure*}

\section{Introduction}
\label{sec:intro}

KV-cache compression helps scale autoregressive decoding by reducing the cost
of attending to a growing history~\citep{pope2023efficiently,gholami2024ai}.
MQA and GQA reduce the number of stored KV heads~\citep{shazeer2019fast,ainslie2023gqa};
Multi-head Latent Attention (MLA), used by DeepSeek-V2/V3, compresses keys and
values jointly~\citep{liu2024deepseek,liu2024deepseekv3}.
Cache size alone does not determine decoding speed: the execution path also
determines computation per cached byte and whether the cache partitions across GPUs.

MLA's MQA-absorb path folds KV up-projections into query/output computations
so attention reads the compact latent. The canonical DeepSeek configuration
at TP$=1$ gives approximately $242$ FLOPs/byte, near H800's BF16 ridge of $295$ but above
H20-3e's $31$~(\citealp{williams2009roofline}; \mbox{\citealp{zadourihardware}}).
Head-wise TP reduces query-side work while replicating the latent; multiple
queries per step increase intensity. The operating regime thus depends on
hardware and parallelism. MLA also permits head expansion, but its head-specific
up-projections do not directly supply shared KV groups for standard GQA.

\begin{figure*}[t]
  \centering
  \begin{subfigure}[b]{0.49\linewidth}
    \centering
    \includegraphics[width=\linewidth,trim=150bp 24bp 0bp 23bp,clip]{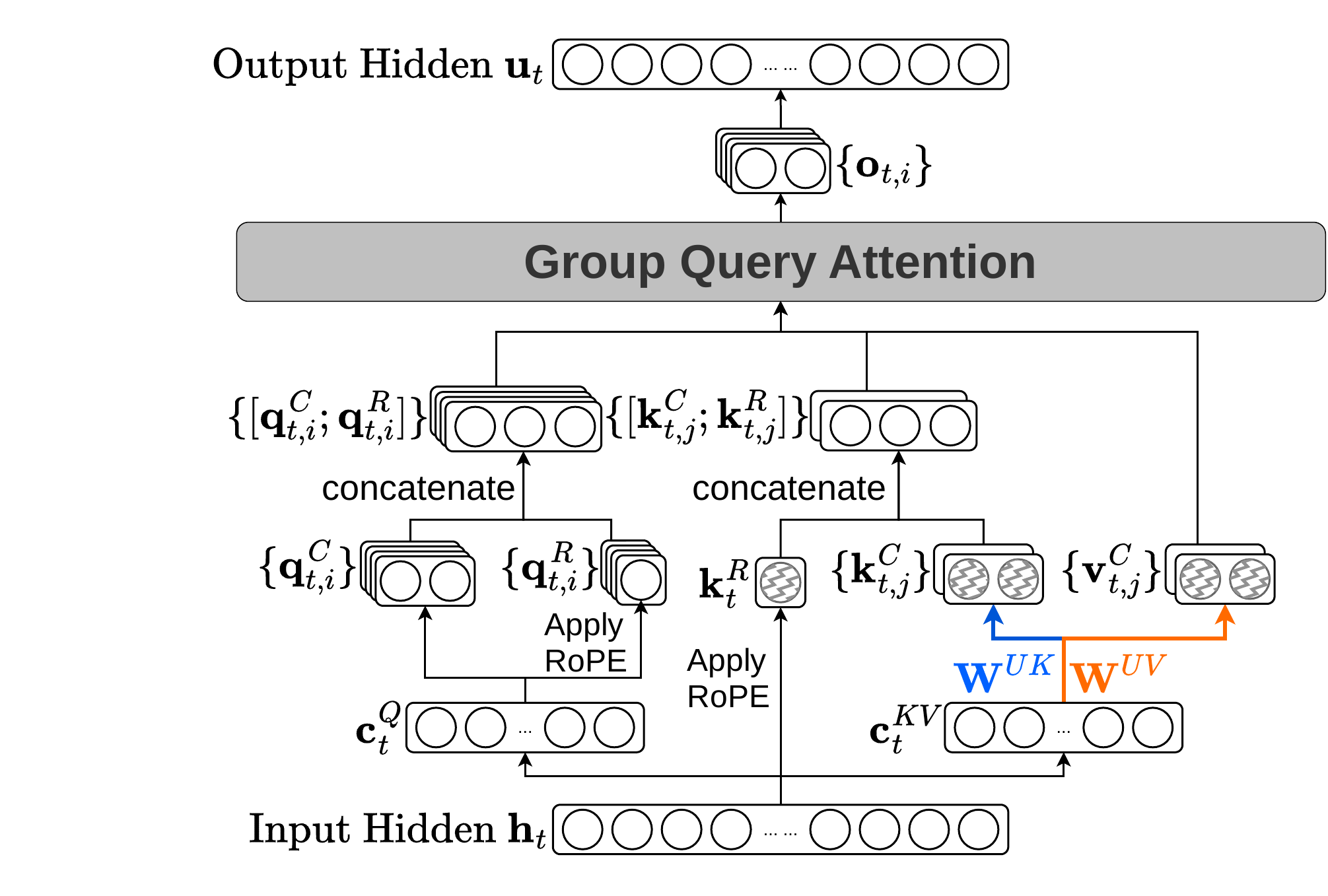}
    \caption{GQA path of GQLA.}
    \label{fig:gqa_mode}
  \end{subfigure}
  \hfill
  \begin{subfigure}[b]{0.49\linewidth}
    \centering
    \includegraphics[width=\linewidth,trim=110bp 24bp 0bp 23bp,clip]{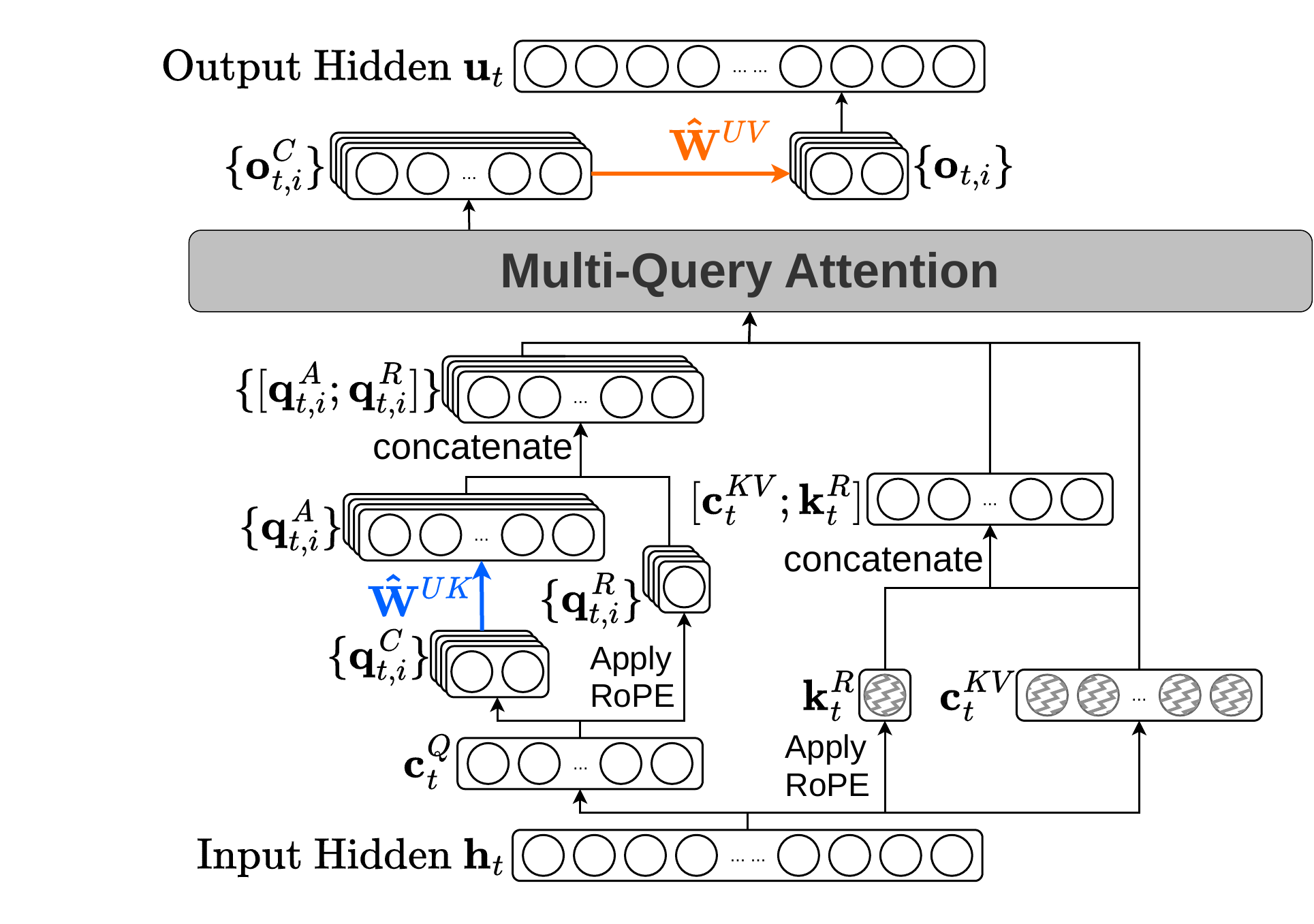}
    \caption{MQA-absorb path of GQLA.}
    \label{fig:mqa_mode}
  \end{subfigure}
\caption{Equivalent GQLA paths. \textbf{Left:} materialize $g$ K/V groups
for a GQA kernel. \textbf{Right:} absorb group up-projections into query/output
computations to attend to the shared latent. Cache traffic, intensity, and TP
partitioning differ, affecting hardware preference (Figure~\ref{fig:roofline}).}
  \label{fig:gqla_modes}
\end{figure*}

\textbf{Group-Query Latent Attention (GQLA)} makes MLA's expanded form practical
for grouped decoding (Figures~\ref{fig:mha_gqa_mqa_mla_gqla} and~\ref{fig:gqla_modes}).
Sharing up-projections within $g$ query-head groups preserves MLA's joint latent
and compact MQA-absorb path. Both paths compute the same function in exact
arithmetic but differ in intensity and cache partitioning. In our deployment,
sharded GQA uses $640$ bytes/token/rank at TP$=8$ versus MQA-absorb's $1152$,
despite its larger cache at TP$=1$.

We first test the modeling constraint imposed by group sharing. With eight
query heads per group, GQLA's terminal training losses are close to MLA's across
the tested model scales; the gap at the largest scale is $0.0041$ with matched tokens.
To access existing checkpoints, \textbf{TransGQLA} extends
TransMLA~\citep{meng2026transmla} by preserving source GQA groups and adds an
MLA route based on per-group activation PCA. Similarity grouping and NLL-weighted PCA refine initialization.
Unlike exact path selection, conversion is approximate: DeepSeek-V3.1-Base
retains the source's eight-task average within $0.05$ percentage points despite
higher perplexity; GQA-source models lose more quality and benefit from training.

Finally, we fix model quality and compare execution paths. Kernels favor
MQA-absorb on H800 and GQA on H20-3e at TP$=1$; GQA wins on both at TP$=8$.
On eight H20-3e GPUs, the converted 671B model's GQA path raises throughput by
$22.42$--$34.97\%$ at matched batches over 8K--64K inputs. Independently tuning
batches to each path's KV capacity at 16K gives $50.07\%$, combining execution
and capacity gains.

Our contributions are:
\begin{enumerate}[label=(\roman*),noitemsep,topsep=2pt,leftmargin=*]
\item \textbf{A joint latent with a shardable expanded cache.} Group-shared
up-projections support equivalent MQA-absorb and GQA execution using existing
kernels (\S\ref{sec:gqla}).
\item \textbf{Training and conversion evidence.} We test pretraining, conversion,
and recovery through continued pretraining (CPT) and supervised fine-tuning
(SFT) (\S\ref{sec:scaling}--\ref{subsec:conversion}).
\item \textbf{Analysis and deployment.} Roofline analysis predicts path preference
across hardware and TP; kernel and full-model measurements test the gains
(\S\ref{sec:roofline}--\ref{sec:exper}).

\end{enumerate}

\section{Related Work}
\label{sec:related}

\paragraph{Attention representations and KV-cache cost.}
MQA~\citep{shazeer2019fast} shares one KV head, GQA~\citep{ainslie2023gqa}
shares one per query-head group, and MLA~\citep{liu2024deepseek} uses a joint latent with
decoupled RoPE. GQLA's grouped up-projections support compact absorbed and
expanded forms. Its shared latent distinguishes it structurally from independent
per-group low-rank compression, an unevaluated baseline.
FlashAttention~\citep{dao2022flashattention} and paged or quantized KV caches
address complementary implementation costs.
\paragraph{Hardware-aware decoding.}
Roofline studies relate attention efficiency to compute and memory
traffic~\citep{williams2009roofline,zadourihardware}, while scaling analyses
identify KV traffic and parallelism as major inference costs~\citep{pope2023efficiently,gholami2024ai}.
We study how factorization affects execution: GQLA's two equivalent paths
have relative costs that vary with GPU specifications and TP degree.

\paragraph{Conversion from pretrained attention.}
TransMLA~\citep{meng2026transmla} and MHA2MLA~\citep{ji2025towards} convert
pretrained attention to latent representations. TransGQLA preserves source GQA
groups during TransMLA-style conversion and uses per-group activation PCA for
MLA. Inspired by sensitivity-aware compression~\citep{lecun1989optimal,frantar2023sparsegpt},
its PCA uses teacher negative log-likelihood (NLL) as heuristic token weights,
without computing a loss Hessian. Similarity grouping chooses which MLA heads
share a KV group.
\paragraph{Training scaling laws.}
Language model scaling laws relate loss to model size, data volume,
and training compute~\citep{kaplan2020scaling,hoffmann2022compute}.
We fit power laws with an additive constant~\citep{henighan2020scaling}
to describe MLA/GQLA training-loss trends under a shared recipe;
these fits do not establish compute-optimal allocation or asymptotic exponents.

\section{Methods}
\label{sec:method}

\subsection{Group-Query Latent Attention}
\label{sec:gqla}

\paragraph{Architecture.}
Query heads in each group share K/V up-projections from one joint latent.
Let $g$ be the KV-group count and $m=h_q/g$ the query heads per group.
Head $i\in[1,h_q]$ uses group $j(i)=\lceil i/m\rceil$; $j\in[1,g]$ indexes groups.
We use equal content-key and value widths $d_h$ below; conversion also supports
a distinct value width $d_h^V$, as in GLM-4.7-Flash.
For token embedding $\mathbf{x}_t \in \R^D$, $W^{DKV} \in \R^{r_{kv} \times D}$
produces latent $\mathbf{c}_{t}^{KV}$. The up-projections
$W^{UK}, W^{UV} \in \R^{g d_h \times r_{kv}}$ expand it into $g$ K/V groups
of width $d_h$, matching a $g$-group GQA cache; $W^{DQ}, W^{UQ}$ analogously
produce $h_q$ query heads. MLA's decoupled RoPE supplies per-head queries
$\mathbf{q}_{t,i}^R \in \R^{d_h^R}$ via $W^{QR}$ and a shared key
$\mathbf{k}_t^R \in \R^{d_h^R}$ via $W^{KR}$:
\begin{align}
\mathbf{c}_{t}^{Q} &= W^{DQ} \mathbf{x}_{t}, \;\; \mathbf{c}_{t}^{KV} = W^{DKV} \mathbf{x}_{t}, \nonumber\\
\mathbf{q}_{t}^{C} &= W^{UQ} \mathbf{c}_{t}^{Q}, \;\; \mathbf{k}_{t}^{C} = W^{UK} \mathbf{c}_{t}^{KV}, \nonumber\\
\mathbf{q}_{t}^{R} &= \rope_t(W^{QR} \mathbf{c}_{t}^{Q}), \;\; \mathbf{k}_{t}^{R} = \rope_t(W^{KR} \mathbf{x}_t), \nonumber\\
\mathbf{q}_{t, i} &= [\mathbf{q}_{t, i}^{C}; \mathbf{q}_{t, i}^{R}], \;\; \mathbf{k}_{t, j} = [\mathbf{k}_{t, j}^{C};\mathbf{k}_{t}^{R}].
\end{align}
Without a Q-side latent rank $r_q$ (as in our LLaMA-3-8B and Qwen2.5-7B
checkpoints), $W^{UQ}W^{DQ}=W^Q$ becomes one matrix acting directly on
$\mathbf{x}_t$, as does $W^{QR}$; subsequent computations are unchanged.

\paragraph{Two equivalent decoding paths.}
The GQA path (Eq.~\eqref{eq:mha}) materializes $g$ K/V groups and caches
$2gd_h+d_h^R$ elements/token with an algorithmically shared RoPE key.
MQA-absorb (Eq.~\eqref{eq:mqa}) attends to the joint latent, caching
$r_{kv}+d_h^R$ elements with the same weights.
We select paths at deployment setup; online cache conversion is not evaluated.
The reused GQA kernel physically repeats RoPE across groups, as accounted for
in \S\ref{sec:roofline}.

\paragraph{GQA path.}
Each query head attends to its assigned group:
\begin{align}
&\mathbf{v}_{t}^{C} = W^{UV} \mathbf{c}_{t}^{KV}, \nonumber\\
&a_{t,s,i}=\softmax_s\!\left(\tfrac{\mathbf{q}_{t,i}^{\top}\mathbf{k}_{s,j(i)}}{\sqrt{d_h+d_h^R}}\right), \nonumber\\
&\mathbf{o}_{t,i}=\sum_{s=1}^{t} a_{t,s,i}\mathbf{v}_{s,j(i)}^{C}, \nonumber\\
&\mathbf{y}_{t} = W^{O} [\mathbf{o}_{t, 1};\dots;\mathbf{o}_{t, h_q}]. \label{eq:mha}
\end{align}
Only per-group $\mathbf{k}^{C}/\mathbf{v}^{C}$ are routed; the single-headed
$\mathbf{k}^{R}$ is shared algorithmically but repeated across KV groups in
the deployed dense GQA cache.

\paragraph{MQA-absorb path.}
\begin{align}
&\hat{W}^{UK}_i=W^{UK}_{j(i)},\qquad \hat{W}^{UV}_i=W^{UV}_{j(i)}, \nonumber\\
&\mathbf{\hat{q}}_{t, i} = [(\hat{W}^{UK}_i)^{\!\top}\mathbf{q}_{t, i}^{C}; \mathbf{q}_{t, i}^{R}], \;\; \mathbf{\hat{k}}_{t} = [\mathbf{c}_{t}^{KV}; \mathbf{k}_{t}^{R}], \;\; \mathbf{\hat{v}}_{t} = \mathbf{c}_{t}^{KV}, \nonumber\\
&\mathbf{\hat{o}}_{t, i} = \sum_{s=1}^{t} \softmax_s\!\left(\tfrac{\mathbf{\hat{q}}_{t, i}^{\top} \mathbf{\hat{k}}_s}{\sqrt{d_h + d_h^{R}}}\right) \mathbf{\hat{v}}_s, \;\; \mathbf{o}_{t, i} = \hat{W}^{UV}_i\mathbf{\hat{o}}_{t, i}, \nonumber\\
&\mathbf{y}_{t} = W^{O} [\mathbf{o}_{t, 1};\dots;\mathbf{o}_{t, h_q}], \label{eq:mqa}
\end{align}
Here $\hat{W}^{UK}_i, \hat{W}^{UV}_i \in \R^{d_h \times r_{kv}}$ are
head $i$'s assigned group blocks; this indexing requires no duplicate storage.
Equivalence follows from $\mathbf{q}^{\top}W\mathbf{c}=(W^{\top}\mathbf{q})^{\top}\mathbf{c}$
and linearity of the value projection through the weighted sum.
With unchanged RoPE and softmax scale, both paths yield identical probabilities
and outputs in exact arithmetic for a fixed GQLA model. Checkpoint conversion
can introduce approximation error.

\subsection{TransGQLA}
\label{sec:transgqla}

\textbf{TransGQLA} converts pretrained GQA and MLA checkpoints using calibration
data without gradient updates. Both routes produce group-indexed up-projections
supporting the two paths of \S\ref{sec:gqla}. The steps below are derived and
specified in full in Appendix~\ref{app:transgqla}.

\subsubsection{From GQA to GQLA}
\label{sec:gqa-to-gqla}

We adapt TransMLA~\citep{meng2026transmla} in three steps.
\textbf{(1) Preserve KV groups:} merge $g$ KV heads into a joint latent, keeping
$W^{UK}, W^{UV}$ group-indexed. Omitting TransMLA's query-head replication
preserves GQA execution and group-axis tensor parallelism.
\textbf{(2) Decouple RoPE:} apply RoRoPE and frequency-aware PCA (FreqFold) to align the RoPE bases and separate positional from non-positional coordinates.
\textbf{(3) Compress the joint latent:} balance the activation norms of $K_{\mathrm{NoPE}}$ and $V$, then jointly compress them to rank $r_{kv}$, retaining a shared $d_h^R$-dimensional RoPE key. The MQA-absorb path thus caches $r_{kv}+d_h^R$ elements per token. Full details appear in Appendix~\ref{app:gqa-to-gqla}.

\paragraph{NLL-weighted PCA.}
For FreqFold and joint K/V compression, we weight activation second moments
by the frozen teacher's per-token next-token NLL, normalized to unit mean per
batch. This loss-weighting heuristic emphasizes difficult calibration tokens;
it does not estimate a loss Hessian. It also applies to MLA PCA (Appendix~\ref{app:mla-to-gqla}). GQA conversion
compresses a newly constructed joint latent; MLA conversion preserves the
latent and approximates head-specific up-projections with group-shared factors.
Each route must therefore be evaluated against its own source.

\subsubsection{From MLA to GQLA}
\label{sec:mla-to-gqla}

MLA conversion preserves the joint latent and reorganizes its head-indexed
up-projections. \textbf{(1) Group similar heads:} score affinity from calibration
K/V covariances, greedily form $g$ balanced groups, and consistently permute
projections. \textbf{(2) Factor each group:} apply separate K-NoPE and V activation
PCA, optionally NLL-weighted, at ranks $r^K=d_h$ and $r^V=d_h^V$.
\textbf{(3) Absorb the factors:} fold per-head reconstruction blocks into
$W^{UQ}$ and $W^O$ without changing their shapes. K/V up-projection rows shrink
from $h_q(d_h+d_h^V)$ to $g(d_h+d_h^V)$; absorption adds no approximation
beyond PCA truncation. Both standard GQA and compact MQA-absorb remain available.
Appendix~\ref{app:mla-to-gqla} details the affinity score, factorization,
absorption equations, and deployment.

\section{Roofline Analysis}
\label{sec:roofline}

\begin{figure*}[t]
  \centering
  \begin{subfigure}[t]{0.49\linewidth}
    \centering
    \includegraphics[width=\linewidth]{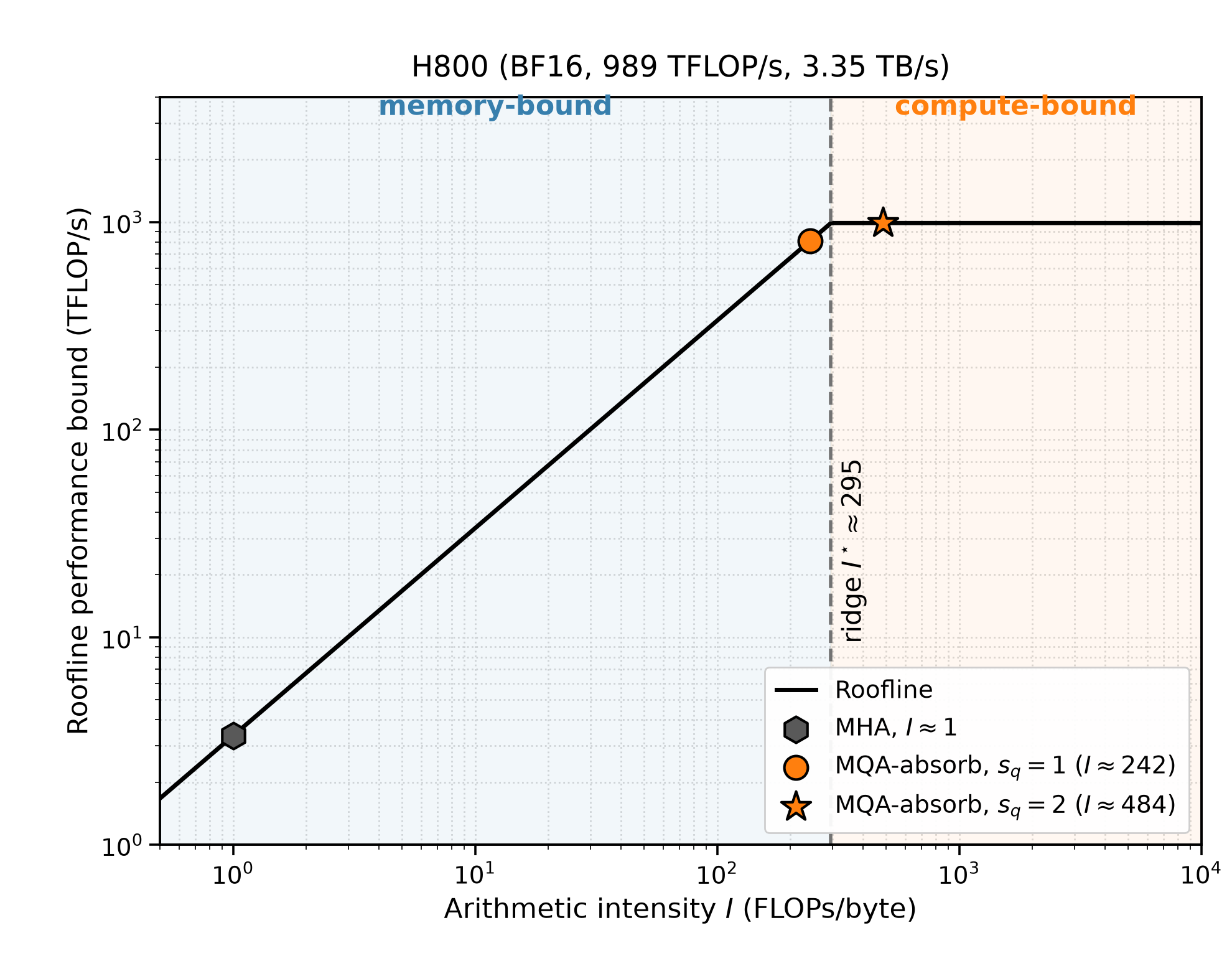}
  \end{subfigure}\hfill
  \begin{subfigure}[t]{0.49\linewidth}
    \centering
    \includegraphics[width=\linewidth]{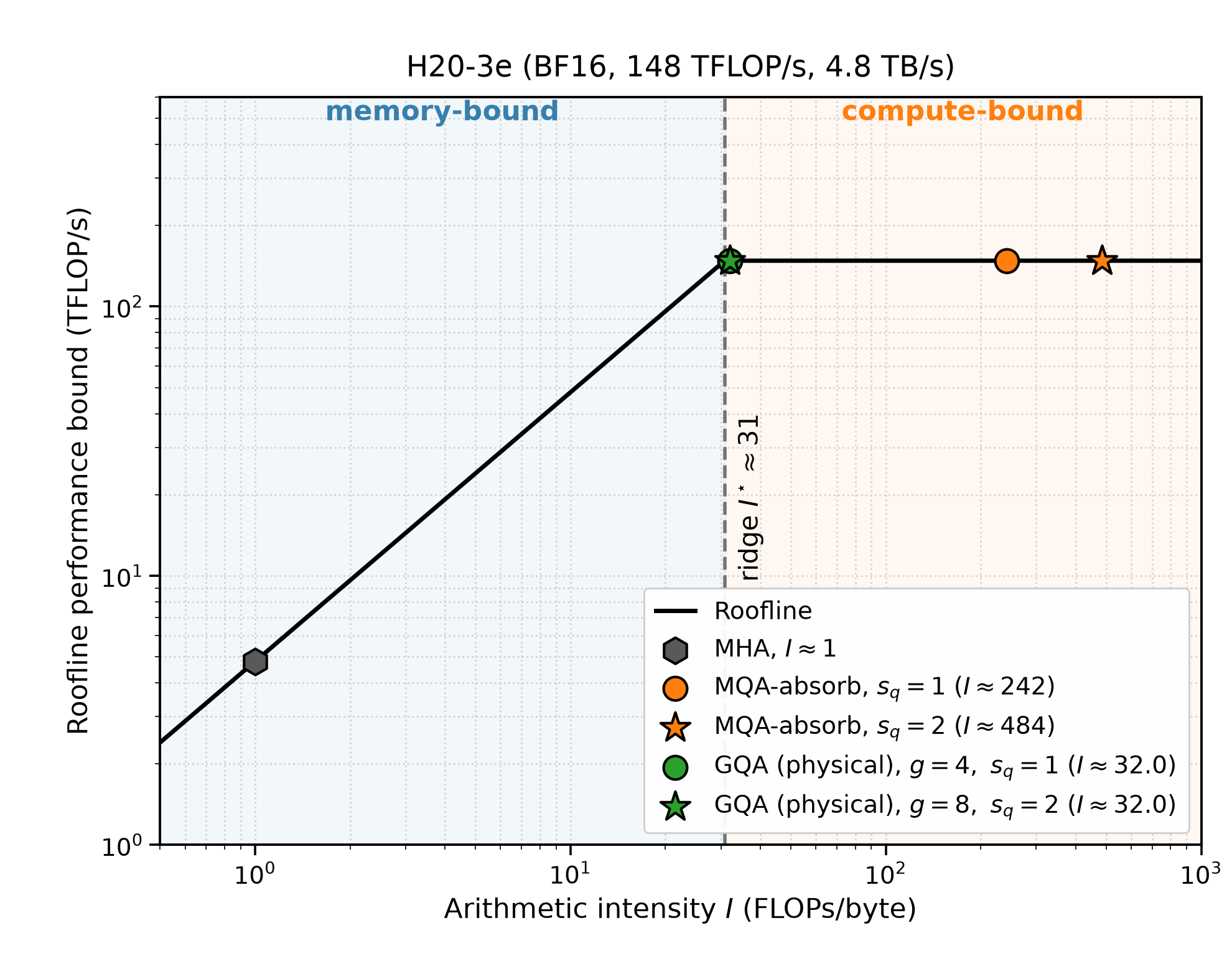}
  \end{subfigure}
  \caption{\small Analytical BF16 Rooflines on H800 (left) and H20-3e (right).
  Solid: throughput bound; dashed: ridge $I^{\star}$.
  MQA-absorb at $s_q=1$ approaches the H800 ridge; on H20-3e,
  GQA at $(g,s_q)\in\{(8,2),(4,1)\}$ reaches $I=32$ under the
  deployed physical cache layout, close to the ridge of $31$ FLOPs/byte.}
  \label{fig:roofline}
\end{figure*}

The Roofline model~\citep{williams2009roofline} bounds throughput by
$\min(I\cdot\mathrm{BW},\mathrm{FLOPs}_{\max})$, with arithmetic intensity
$I=F/B$ and ridge $I^\star=\mathrm{FLOPs}_{\max}/\mathrm{BW}$.
The hardware parameters in Table~\ref{tab:hardware} give
$I^\star\!\approx\!295$ for H800 and $31$ for H20-3e,
corresponding to the dashed lines in Figure~\ref{fig:roofline}.
We analyze the canonical DeepSeek-V3.1 configuration
$(h_q,g,d_h,d_h^R,r_{kv})=(128,8,128,64,512)$, with $s_q$ queries per
step ($s_q=1$ for decoding, $s_q=2$ for one multi-token prediction (MTP) head).
The model includes leading attention cache reads and matrix products, omitting
launch overhead, communication, and other layers. \S\ref{sec:exper} measures
realized performance.

\begin{table}[t]
\centering
\small
\setlength{\tabcolsep}{4pt}
\begin{tabular}{lccc}
\toprule
GPU & BF16 (TFLOP/s) & HBM (TB/s) & Ridge (FLOPs/byte) \\
\midrule
H800 & $989$ & $3.35$ & $\approx 295$ \\
H20-3e & $148$ & $4.8$ & $\approx 31$ \\
\bottomrule
\end{tabular}
\caption{Hardware parameters for Figure~\ref{fig:roofline}.
Dense BF16 peaks (no 2:4 sparsity) are vendor-reported;
$I^\star=\mathrm{FLOPs}_{\max}/\mathrm{BW}$.}
\label{tab:hardware}
\end{table}

\paragraph{Cache traffic and arithmetic intensity.}
\label{subsec:gqla-roofline}
MQA-absorb caches $2(r_{kv}+d_h^R)=1152$ bytes/token/layer in BF16.
The deployed GQA kernel repeats the shared RoPE key across groups,
using $2g(2d_h+d_h^R)=5120$ bytes; the algorithmic shared-RoPE layout
would use $4224$ bytes. Assuming each step reads the $L$ cached tokens
once and reuses them across queries, the two intensities are
\begin{equation}
I_{\mathrm{MQA}}=\frac{h_qs_q(2r_{kv}+d_h^R)}{r_{kv}+d_h^R},
\qquad I_{\mathrm{GQA,physical}}=\frac{h_qs_q}{g}.
\label{eq:roofline-summary}
\end{equation}
Full byte/FLOP derivations and operating points are in
Appendix~\ref{app:roofline-details}; additional hardware and cache-layout
comparisons are in Appendix~\ref{app:h20-comparison}.

\paragraph{Path selection and MTP.}
At TP$=1$, MQA-absorb has $I\approx242$ for $s_q=1$ and $484$ for $s_q=2$.
The first approaches H800's ridge; the second is compute-bound.
At $L=8192$, predicted step time rises from $2.82$ to $4.61\,\mu$s,
limiting the query-throughput gain to $1.22\times$; both points are
compute-bound on H20-3e. GQA has $I=16$ for $s_q=1$ and $32$ for $s_q=2$.
The latter approaches H20-3e's ridge (Figure~\ref{fig:roofline}), predicting
$220.5$K query positions/s versus MQA-absorb's $64.9$K ($3.4\times$).
The model favors $(g,s_q)=(8,2)$ for KV sharding up to TP$=8$;
$(4,1)$ has the same intensity and half the physical cache, with sharding
up to TP$=4$. Both limits assume no KV replication.

\paragraph{Tensor parallelism.}
For $p$ dividing $g$, GQA shards query heads and KV groups together, preserving
intensity and dividing per-rank cache traffic by $p$. MQA-absorb replicates
the latent, dividing intensity by $p$. At TP$=8$, physical GQA uses $640$
bytes/token/rank versus MQA-absorb's $1152$, giving $1.8\times$ the token capacity
at the same KV-memory budget. Thus the smaller unsharded cache becomes larger
per rank, making path preference depend on TP and GPU type.

\paragraph{When expansion reduces per-rank memory.}
For TP degree $p$ dividing $g$, the physical per-rank cache ratio is
\begin{equation}
\frac{B_{\mathrm{GQA}}^{\mathrm{rank}}}{B_{\mathrm{MQA}}^{\mathrm{rank}}}
=\frac{g(2d_h+d_h^R)}{p(r_{kv}+d_h^R)}.
\label{eq:rank-cache-ratio}
\end{equation}
GQA uses less memory per rank when $p>g(2d_h+d_h^R)/(r_{kv}+d_h^R)$.
For the canonical configuration this threshold is approximately $4.44$:
the GQA cache is still larger at TP$=4$ ($1280$ versus $1152$ bytes/token),
but smaller at TP$=8$ ($640$ versus $1152$).
Thus expansion saves per-rank memory only when enough groups are sharded.

Increasing $g$ permits more KV shards but enlarges the unsharded cache;
increasing $s_q$ reuses cache reads but raises intensity. For a fixed $g$, both paths remain
available; measurements at the intended batch, context, and TP determine preference.

\paragraph{Interpreting the ideal speedup.}
The model separates memory and arithmetic savings. At TP$=8$ on H20-3e,
both paths are memory-bound for $s_q=1$, giving predicted speedup
$1152/640=1.8$ from the cache ratio. For $s_q=2$, both are compute-bound,
giving the FLOP ratio $(2r_{kv}+d_h^R)/(2d_h+d_h^R)=3.4$.
Realizing these attention-only speedups depends on kernel efficiency;
they exclude other layers and serving communication.

\section{Experiments}
\label{sec:exper}

We test group sharing through pretraining, evaluate conversion and recovery,
then compare execution paths at fixed quality. Appendix~\ref{app:exper-details}
gives configurations and protocols.

\subsection{Scaling with Pretraining Compute}
\label{sec:scaling}

\paragraph{Setup.}
We test the cost of group sharing by pretraining MLA and GQLA from scratch.
GQLA 1:4 and 1:8 use $m=4$ and $m=8$ query heads per KV group; $g=h_q/m$. MLA and GQLA 1:8 are evaluated at six model scales;
GQLA 1:4 uses the four smaller scales (Appendix~\ref{app:scaling-details}).
At each size, all methods use the same data, token budget, optimizer, and
learning rate schedule. Terminal training loss averages the final 100 iterations.
This measures training fit; it does not measure held-out or downstream quality.

\begin{figure}[htbp]
  \centering
  \includegraphics[width=\linewidth]{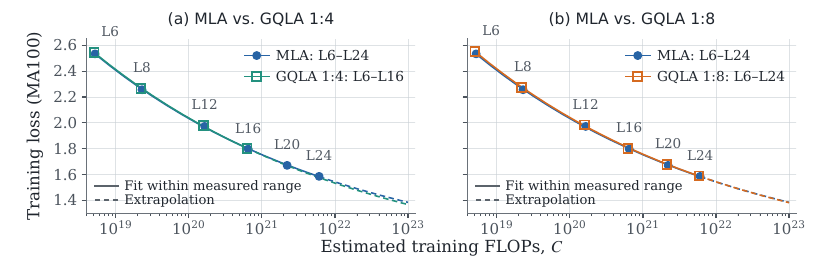}
\caption{MLA/GQLA scaling under a shared pretraining recipe. Markers: terminal
loss averaged over the final 100 iterations. Solid curves: fits within observed
compute ranges; dashed curves: extrapolation to $10^{23}$ FLOPs.}
  \label{fig:scaling_ladder}
\end{figure}

\paragraph{Scaling fit.}
We model the loss--compute relationship using a power law with an additive
constant~\citep{henighan2020scaling}:
\begin{equation}
L(C)=L_{\infty}+A\left(\frac{C}{10^{21}\,\mathrm{FLOPs}}\right)^{-\alpha}.
\label{eq:scaling-law}
\end{equation}
Cumulative training compute $C$ is estimated from processed tokens and each
attention architecture's FLOP cost. Each method's three parameters are fitted
independently by unweighted nonlinear least squares in loss space.

\paragraph{Results.}
Both GQLA variants track MLA's compute--loss curve over the tested scales
(Figure~\ref{fig:scaling_ladder}). At matched token budgets, GQLA 1:8's loss
gap to MLA narrows from $0.0184$ at the smallest scale to $0.0041$ at the largest;
the descriptive fits give $\alpha=0.123$ and $0.124$, respectively. GQLA 1:4
follows a similar trend, with a $0.0032$ gap at its largest tested scale and matched tokens.
Thus group sharing gives similar terminal training losses in the measured range.
Each three-parameter fit uses six endpoints for MLA and 1:8, or four for 1:4.
Similar exponents and dashed extrapolations do not establish equal asymptotic
laws. Matched-token gaps are distinct from compute-based comparisons.

\subsection{Model Conversion and Recovery}
\label{subsec:conversion}

\paragraph{Setup.}
We convert LLaMA-3-8B ($g=8$) and Qwen2.5-7B ($g=4$) from GQA, and
GLM-4.7-Flash ($g=4$) and DeepSeek-V3.1-Base ($g=8$) from MLA.
Calibration uses $128$ WikiText-2 samples with length $256$ for GQA sources
and $512$ for MLA sources, without gradient updates.
The GQA conversions cache $576$ latent elements/token on MQA-absorb
($28.125\%$ of LLaMA's source cache and $56.25\%$ of Qwen's);
MLA conversions retain the source latent.
We report WikiText-2 perplexity (PPL) and the unweighted mean of eight zero-shot tasks:
PIQA, Winogrande, ARC-e, ARC-c, BoolQ, MMLU, HellaSwag, and OpenBookQA.
PPL uses length $2048$, except $1024$ for DeepSeek-V3.1.
Recovery uses $30$B CPT tokens for LLaMA, $1$B for Qwen, and $12$B for GLM,
each followed by approximately $47$M SFT tokens, with one seed per run.

\begin{table*}[t]
\centering
\small
\setlength{\tabcolsep}{4pt}
\caption{Source, calibration-only, and CPT/SFT quality. WikiText-2 PPL uses length
$2048$ ($1024$ for DeepSeek-V3.1); Avg is unweighted eight-task accuracy (\%).
Here $g$ is the target group count. NLL PCA weights by token NLL;
grouping uses head similarity (MLA sources only).
Each block lists the source, conversion with ablations, and recovery;
TransMLA is the GQA-route baseline. Bold scores mark the best calibration-only
conversion; bold labels select recovery/deployment initializations.
DeepSeek uses $g=8$ (16 heads/group), reducing expanded K/V but retaining the latent cache.}
\resizebox{\textwidth}{!}{%
\begin{tabular}{lcccccccccc}
\toprule
\textbf{Variant} & \textbf{PPL}$\downarrow$ & \textbf{PIQA} & \textbf{WG} & \textbf{ARC-e} & \textbf{ARC-c} & \textbf{BoolQ} & \textbf{MMLU} & \textbf{HS} & \textbf{OBQA} & \textbf{Avg} \\
\midrule
\multicolumn{11}{c}{\emph{LLaMA-3-8B} ($h_q\!=\!32$, $g\!=\!8$, $d_h\!=\!128$, $d_h^R\!=\!64$, $r_{kv}\!=\!512$)} \\
\midrule
\rowcolor{gray!10}Source GQA
& $6.14$ & $78.67$ & $73.09$ & $80.77$ & $51.88$
& $81.90$ & $62.07$ & $79.21$ & $44.60$ & $69.02$ \\
TransMLA
& $25.77$ & $73.45$ & $66.38$ & $67.97$ & $38.05$
& $58.87$ & $37.39$ & $66.92$ & $39.20$ & $56.03$ \\
GQA$\to$GQLA
& $25.87$ & $73.88$ & $66.54$ & $67.80$ & $\mathbf{38.23}$
& $58.87$ & $37.25$ & $66.99$ & $\mathbf{39.60}$ & $56.14$ \\
\textbf{\;+ NLL PCA}
& $\mathbf{18.57}$ & $\mathbf{74.86}$ & $\mathbf{67.64}$
& $\mathbf{70.45}$ & $37.46$ & $\mathbf{60.80}$
& $\mathbf{41.36}$ & $\mathbf{67.13}$ & $35.80$ & $\mathbf{56.94}$ \\
\;+ 30B CPT + $\sim$47M SFT
& $7.06$ & $81.07$ & $73.24$ & $81.94$ & $50.09$
& $84.80$ & $57.61$ & $78.45$ & $45.00$ & $69.02$ \\
\midrule
\multicolumn{11}{c}{\emph{Qwen2.5-7B} ($h_q\!=\!28$, $g\!=\!4$, $d_h\!=\!128$, $d_h^R\!=\!64$, $r_{kv}\!=\!512$)} \\
\midrule
\rowcolor{gray!10}Source GQA
& $6.85$ & $78.73$ & $73.09$ & $80.51$ & $48.29$
& $84.62$ & $71.81$ & $78.84$ & $47.20$ & $70.39$ \\
TransMLA
& $8.41$ & $\mathbf{78.29}$ & $\mathbf{69.46}$
& $76.18$ & $44.62$ & $81.53$ & $65.89$
& $78.35$ & $44.40$ & $\mathbf{67.34}$ \\
GQA$\to$GQLA
& $8.41$ & $78.07$ & $69.30$ & $\mathbf{76.26}$
& $\mathbf{44.71}$ & $\mathbf{81.56}$ & $65.88$
& $\mathbf{78.39}$ & $44.40$ & $67.32$ \\
\textbf{\;+ NLL PCA}
& $\mathbf{8.25}$ & $78.07$ & $68.59$ & $75.93$ & $43.86$
& $81.22$ & $\mathbf{66.18}$ & $77.92$
& $\mathbf{45.60}$ & $67.17$ \\
\;+ 1B CPT + $\sim$47M SFT
& $7.29$ & $80.14$ & $71.74$ & $82.74$ & $53.58$
& $85.17$ & $68.96$ & $76.51$ & $46.00$ & $70.61$ \\
\midrule
\multicolumn{11}{c}{\emph{GLM-4.7-Flash}
($h_q\!=\!20$, $g\!=\!4$, $d_h\!=\!192$, $d_h^R\!=\!64$,
$d_h^V\!=\!256$, $r_{kv}\!=\!512$)} \\
\midrule
\rowcolor{gray!10}Source MLA
& $11.39$ & $79.82$ & $72.61$ & $82.20$ & $55.63$
& $88.38$ & $70.83$ & $80.13$ & $44.60$ & $71.77$ \\
MLA$\to$GQLA
& $10.61$ & $77.42$ & $68.90$ & $77.57$ & $49.40$
& $80.64$ & $61.69$ & $73.40$ & $41.60$ & $66.33$ \\
\;+ NLL PCA
& $\mathbf{10.41}$ & $77.91$ & $\mathbf{70.24}$
& $78.16$ & $49.66$ & $82.78$ & $62.43$
& $73.76$ & $42.60$ & $67.19$ \\
\;+ grouping
& $10.84$ & $77.58$ & $69.38$ & $\mathbf{80.68}$
& $51.54$ & $\mathbf{84.77}$ & $\mathbf{63.74}$
& $74.00$ & $\mathbf{43.80}$ & $68.19$ \\
\textbf{\;+ grouping + NLL PCA}
& $11.10$ & $\mathbf{78.40}$ & $69.38$ & $80.39$
& $\mathbf{52.56}$ & $83.67$ & $63.20$
& $\mathbf{74.99}$ & $43.20$ & $\mathbf{68.22}$ \\
\;+ 12B CPT + $\sim$47M SFT
& $6.73$ & $81.01$ & $74.51$ & $83.88$ & $56.06$
& $85.69$ & $66.64$ & $78.54$ & $46.80$ & $71.64$ \\
\midrule
\multicolumn{11}{c}{\emph{DeepSeek-V3.1-Base} ($h_q\!=\!128$, $g\!=\!8$, $d_h\!=\!128$, $d_h^R\!=\!64$, $r_{kv}\!=\!512$)} \\
\midrule
\rowcolor{gray!10}Source MLA
& $3.21$ & $83.35$ & $81.45$ & $86.87$ & $63.48$
& $88.47$ & $85.12$ & $86.56$ & $47.40$ & $77.84$ \\
MLA$\to$GQLA
& $8.19$ & $\mathbf{83.46}$ & $81.06$ & $86.74$
& $\mathbf{63.74}$ & $88.41$ & $\mathbf{85.11}$
& $86.60$ & $\mathbf{47.60}$ & $77.84$ \\
\;+ NLL PCA
& $8.32$ & $83.19$ & $80.98$ & $86.66$ & $63.65$
& $\mathbf{88.56}$ & $85.09$ & $\mathbf{86.68}$
& $47.40$ & $77.78$ \\
\;+ grouping
& $6.21$ & $83.19$ & $\mathbf{81.53}$ & $86.74$
& $\mathbf{63.74}$ & $88.44$ & $85.05$
& $86.60$ & $\mathbf{47.60}$ & $\mathbf{77.86}$ \\
\textbf{\;+ grouping + NLL PCA}
& $\mathbf{6.05}$ & $83.41$ & $81.37$ & $\mathbf{86.83}$
& $63.31$ & $88.53$ & $85.07$
& $86.61$ & $47.20$ & $77.79$ \\
\bottomrule
\end{tabular}}
\label{tab:conversion}
\end{table*}

\paragraph{Conversion quality and ablations.}
GQA sources incur joint-latent compression loss (Table~\ref{tab:conversion}).
Plain conversion closely matches TransMLA; NLL weighting reduces LLaMA PPL
by $28.2\%$ and Qwen PPL by $1.9\%$. LLaMA still drops from $69.02$ to $56.94$ Avg.
For MLA sources, grouping approximates the up-projections. On GLM, grouping
raises Avg from $66.33$ to $68.19$; adding weighting gives $68.22$ versus the
source's $71.77$. Weighting alone gives the lowest PPL ($10.41$), showing
that the best configuration depends on the metric.

DeepSeek illustrates why both metrics matter. Grouping lowers PPL from
$8.19$ to $6.21$; weighting gives $6.05$, a $26\%$ improvement over plain
conversion but still above the source's $3.21$. Avg remains close
($77.79$ versus $77.84$). This is a useful deployment tradeoff without further
training, not lossless language modeling. Here $g=8$ and $m=16$; the
$m=8$ pretraining result does not establish quality at this stronger sharing ratio.

\paragraph{What the refinements improve.}
On GLM, grouping adds $1.86$ Avg points without weighting and $1.03$ with it;
weighting adds $0.86$ points without grouping and $0.03$ after grouping.
On Qwen, weighting lowers PPL from $8.41$ to $8.25$ while Avg changes from
$67.32$ to $67.17$. Likelihood gains need not improve task accuracy, and the combined refinements'
gains are not additive.

\paragraph{Recovery.}
CPT/SFT yields Avg $69.02$ and PPL $7.06$ on LLaMA, $70.61$ and $7.29$ on
Qwen, and $71.64$ and $6.73$ on GLM. All three averages lie within $0.22$
percentage points of their sources, although recovery budgets differ substantially.
Average recovery does not imply uniform task recovery: LLaMA matches the
source Avg of $69.02$ but scores $57.61$ versus $62.07$ on MMLU.
Table~\ref{tab:conversion} reports all tasks.

\subsection{Kernel-Level Validation}
\label{subsec:kernel-validation}

We compare FA3-MLA~\citep{shah2024flashattention3} (MQA-absorb)
with HPC-Ops\footnote{\url{https://github.com/Tencent/hpc-ops}} (GQA)
at $B=128$, $L=8192$, BF16, page size $64$,
and cold L2 after initialization and correctness checks.
TP is simulated through per-rank shapes on one GPU, excluding communication
and other model components. Throughput counts query positions per GPU,
including both at $s_q=2$, without multiplying by TP.
Gains at $s_q=2$ measure cache-read reuse across queries; generated-token
throughput would additionally depend on MTP acceptance and scheduling,
which this benchmark does not evaluate.

\begin{table}[t]
\centering
\small
\setlength{\tabcolsep}{6pt}
\caption{Measured kernel throughput (K query positions/s).
MQA-absorb uses FA3-MLA; GQA uses HPC-Ops with $g=8$.
TP columns use rank-local shapes on one GPU, excluding communication.
The complete TP sweep and analytical comparison are in Table~\ref{tab:roofline}.}
\begin{tabular}{lcrrrr}
\toprule
& & \multicolumn{2}{c}{TP$=1$} & \multicolumn{2}{c}{TP$=8$} \\
GPU & $s_q$ & MQA-absorb & GQA & MQA-absorb & GQA \\
\midrule
H800 & 1 & 261.3 & 74.7 & 312.7 & 558.2 \\
     & 2 & 272.3 & 148.0 & 619.7 & 1106.0 \\
H20-3e & 1 & 47.4 & 100.6 & 95.4 & 742.0 \\
       & 2 & 54.0 & 145.2 & 190.1 & 1045.0 \\
\bottomrule
\end{tabular}
\label{tab:kernel-summary}
\end{table}

\paragraph{TP can reverse path preference.}
At TP$=8$, GQA is $7.78\times$ faster on H20-3e for $s_q=1$ and
$5.50\times$ for $s_q=2$ (Table~\ref{tab:kernel-summary}). On H800,
MQA-absorb wins at TP$=1$, but GQA is $1.79\times$ faster at TP$=8$.
For one query, TP$=8$ reduces MQA-absorb's intensity to $30.2$ while GQA
retains $16$ FLOPs/byte; two queries double these to $60.4$ and $32$.
The H20-3e Roofline predicts gains of $1.80\times$ for one query and
$3.40\times$ for two. Larger measured gains also reflect kernel efficiency.
Appendix~\ref{app:kernel-details} gives the full TP sweep and analysis.

\subsection{End-to-End Serving}
\label{subsec:end-to-end-serving}

We serve converted DeepSeek-V3.1-Base ($g=8$) on eight 141\,GB H20-3e GPUs
(TP$=8$, PP$=1$). The same checkpoint runs FA3-MLA MQA-absorb
or HPC-Ops GQA, holding quality fixed; Table~\ref{tab:conversion} separately
evaluates conversion quality. Inputs of 8K, 16K, 32K, and 64K use matched resident batches of $64$, $40$,
$20$, and $10$, respectively, with $512$ output tokens, BF16 compute/KV caches,
and no MTP. Each run submits $2B$ requests at concurrency $2B$. We average five runs with
fresh servers, excluding client warmups (Appendix~\ref{app:serving-details}).
Output throughput, time to first token (TTFT), time per output token (TPOT),
and end-to-end latency include serving costs beyond attention. Live KV counts resident tokens in the
physical layout, excluding weights, workspace, and other allocations. Matched
comparisons fix resident batches; capacity tuning uses extra KV capacity.

\begin{table*}[!t]
\centering
\small
\setlength{\tabcolsep}{4pt}
\caption{Serving the same converted DeepSeek-V3.1-Base checkpoint on eight
H20-3e GPUs (TP$=8$, five-run means). All arrows run from MQA-absorb (FA3-MLA)
to GQA (HPC-Ops). Matched rows fix resident batch and client workload;
the final row independently tunes batches to capacity. Latencies give relative
reductions; live KV/GPU assumes $B(L+512)$ tokens across $61$ layers.}
\resizebox{\textwidth}{!}{%
\begin{tabular}{llccccccc}
\toprule
\textbf{Context}
& \textbf{Setting}
& \textbf{$B$}
& \shortstack{\textbf{Live KV/GPU}\\\textbf{(GiB)}}
& \shortstack{\textbf{Output}\\\textbf{(tok/s)}}
& \shortstack{\textbf{Throughput}\\$\uparrow$}
& \textbf{TPOT$\downarrow$}
& \textbf{TTFT$\downarrow$}
& \textbf{E2E$\downarrow$} \\
\midrule
8K
& Matched
& $64\!\to\!64$
& $36.46\!\to\!20.25$
& $330.43\!\to\!\mathbf{405.20}$
& $\mathbf{22.63\%}$
& $\mathbf{21.18\%}$
& $\mathbf{13.53\%}$
& $\mathbf{17.46\%}$ \\

16K
& Matched
& $40\!\to\!40$
& $44.23\!\to\!24.57$
& $143.31\!\to\!\mathbf{193.42}$
& $\mathbf{34.97\%}$
& $\mathbf{30.29\%}$
& $\mathbf{18.13\%}$
& $\mathbf{24.31\%}$ \\

32K
& Matched
& $20\!\to\!20$
& $43.56\!\to\!24.20$
& $61.68\!\to\!\mathbf{80.63}$
& $\mathbf{30.72\%}$
& $\mathbf{28.30\%}$
& $\mathbf{15.77\%}$
& $\mathbf{21.87\%}$ \\

64K
& Matched
& $10\!\to\!10$
& $43.23\!\to\!24.01$
& $24.01\!\to\!\mathbf{29.40}$
& $\mathbf{22.42\%}$
& $\mathbf{23.26\%}$
& $\mathbf{11.66\%}$
& $\mathbf{16.90\%}$ \\
\midrule
16K
& Capacity-tuned
& $40\!\to\!72$
& $44.23\!\to\!44.23$
& $143.31\!\to\!\mathbf{215.05}$
& $\mathbf{50.07\%}$
& -- & -- & -- \\
\bottomrule
\end{tabular}}
\label{tab:serving}
\end{table*}
\paragraph{Faster execution and greater capacity.}
At matched workloads, the GQA path improves output throughput by $22.42$--$34.97\%$
and lowers mean TPOT by $21.18$--$30.29\%$ (Table~\ref{tab:serving}), including
prefill, MoE computation, TP communication, and scheduling.
Sharding reduces live KV/GPU by $44.4\%$, from $44.23$ to $24.57$ GiB at 16K.
Capacity tuning uses this headroom to raise GQA's batch from $40$ to $72$
at the same $44.23$ GiB budget, reaching $215.05$ versus MQA-absorb's $143.31$
output tok/s. This $50.07\%$ gain combines execution and capacity improvements;
the matched-batch gain at 16K is $34.97\%$.

Across the four contexts, matched workloads also lower mean TTFT by
$11.66$--$18.13\%$ and end-to-end latency by $16.90$--$24.31\%$.
These include prefill and scheduling. At 16K, raising GQA's batch from $40$ to $72$ increases throughput
from $193.42$ to $215.05$ tok/s: an additional $11.18\%$ from capacity beyond
the gain at matched load.

\subsection{Limitations}
\label{sec:limitations}
Results apply to the tested checkpoints and implementations. Roofline predictions
use peak specifications; kernel gains also depend on relative FA3-MLA/HPC-Ops
efficiency. Full-model serving covers only H20-3e at TP$=8$, without isolated
TP scaling or communication studies. The $s_q=2$ kernel results do not establish
end-to-end MTP speedup or acceptance rates. Calibration uses one dataset and
$128$ samples; recovery uses one seed. Small accuracy differences are not
statistically established improvements. We do not evaluate independently
compressed KV groups or alternative MLA-source conversion methods.

\section{Conclusion}
\label{sec:conclusion}

GQLA combines a joint latent with group-shared up-projections, enabling
shardable GQA and equivalent MQA-absorb execution across hardware and TP.
Pretraining at $m=8$ yields terminal losses close to MLA's; TransGQLA measures
conversion and recovery separately. On the converted 671B model ($g=8,m=16$),
GQA improves throughput by $22.42$--$34.97\%$ at matched batches on eight
H20-3e GPUs; capacity tuning gives $50.07\%$. These results support multiple
execution forms of a learned function, with conversion quality assessed separately.

\clearpage
\section*{AI Usage Disclosure}

We used OpenAI Codex to assist with language editing and manuscript organization,
clarify the interpretation and limitations of reported results, check mathematical
notation and the algebraic arguments relating the two execution paths, and prepare
code for numerical verification and analytical Roofline figures. During revision,
numerical entries were compared with the existing manuscript, and causal-attention
and PCA-absorption identities were checked numerically, including unequal key and
value widths. The authors are responsible for the final text, claims, code, and
figures, including AI-assisted content. This disclosure follows the
\href{https://iclr.cc/Conferences/2027/AIPolicyForAuthors}{ICLR 2027 AI policy for authors}.

\bibliography{iclr2027_conference}
\bibliographystyle{iclr2027_conference}

\appendix
\section{TransGQLA: Full Conversion Details}
\label{app:transgqla}

Both GQA checkpoints (e.g., LLaMA-3~\citep{grattafiori2024llama}) and MLA checkpoints (DeepSeek-V2/V3~\citep{liu2024deepseek,liu2024deepseekv3}, GLM-4.7-Flash~\citep{glm47}) can be converted to GQLA without pretraining from scratch. We refer to both routes collectively as \textbf{TransGQLA}: \S\ref{app:gqa-to-gqla} extends TransMLA~\citep{meng2026transmla} with a targeted head-merging change that keeps the up-projections group-indexed, while \S\ref{app:mla-to-gqla} recovers the same factorization from a pretrained MLA latent via group-wise PCA on up-projection activations. Both produce weights that expose the two paths of \S\ref{sec:gqla}.

\subsection{From GQA to GQLA}
\label{app:gqa-to-gqla}

This route reuses the entire TransMLA pipeline (head merging, RoRoPE, FreqFold, KV-norm balancing) with one change at head merging.

\paragraph{Merging grouped heads to a latent head.}
TransMLA folds GQA's $g$ KV heads into a latent and \emph{replicates} $W^{UK}, W^{UV}$ across all $h_q$ query heads, so the merged module behaves as MHA. TransGQLA drops the replication: $W^{UK}, W^{UV}$ remain indexed by $j \in [1, g]$, so the merged module is standard GQA identical to the GQA path of \S\ref{sec:gqla}, with MQA-absorb still reachable exactly as in MLA. The expanded GQA form therefore permits group-wise cache sharding, whereas head-wise TP of the absorbed MLA form replicates the shared latent.

The merged GQA attention is re-expressed as
\begin{align}
&\mathbf{q}_{t} = W^{Q} \mathbf{x}_{t}, \;\; \mathbf{c}_{t}^{KV} = [\mathbf{c}_{t}^{K}; \mathbf{c}_{t}^{V}] = W^{DKV} \mathbf{x}_{t},  \nonumber\\
&\mathbf{\hat{q}}_{t, i}^R = \hrope_t \!\left((W^{UK}_{j(i)})^{\!\top}\mathbf{q}_{t, i}\right), \nonumber\\
&\mathbf{\hat{k}}_t^R = \hrope_t(\mathbf{c}_{t}^K), \;\; \mathbf{\hat{v}}_t = \mathbf{c}_{t}^V,  \nonumber\\
&\mathbf{\hat{o}}_{t, i} = \sum_{s=1}^{t} \softmax_s\!\left(\tfrac{\mathbf{\hat{q}}^{R^\top}_{t,i}\mathbf{\hat{k}}_{s}^R}{\sqrt{d_h}}\right) \mathbf{\hat{v}}_{s}, \nonumber\\
&\mathbf{y}_{t} = W^{O}  [W^{UV}_{j(1)}\mathbf{\hat{o}}_{t, 1};\dots;W^{UV}_{j(h_q)}\mathbf{\hat{o}}_{t, h_q}], \label{eq:transgqla}
\end{align}
where $j(i) = \lceil i / (h_q/g) \rceil$ routes head $i$ to its group, $W^{UK}_j = W^{UV}_j \in \R^{d_h \times g d_h}$ is initialized as a sparse identity selecting group $j$ (mirroring GQA's \texttt{repeat\_kv}), and $\hrope$ folds the $g$ identical per-head rotations into one repeating every $d_h$ dimensions. The cache stays at $2 g d_h$ until the subsequent pipeline compresses it.

\paragraph{RoRoPE: aligning per-head RoPE bases.}
After the GQA-preserving merge each of the $g$ KV heads still carries its own RoPE basis, blocking a single shared decoupled key. RoRoPE uses a per-group orthogonal change of basis $R_j \in \R^{d_h \times d_h}$, applied consistently to keys, matching queries, and rotary operators. The query-side change is absorbed into the matching projection slices. Attention scores are preserved while the groups' RoPE bases are aligned. The result is the single $\hrope$ in Eq.~\eqref{eq:transgqla}, which repeats every $d_h$ dimensions of the latent in place of $g$ per-head rotations.

\paragraph{FreqFold: frequency-aware PCA.}
A naive PCA on the merged $g d_h$-dim K/V matrix would mix rotated and non-rotated coordinates and destroy the RoPE pair structure. FreqFold partitions dimensions by RoPE frequency band and applies PCA \emph{within} each band, rotating the RoPE/NoPE coordinates into axis-aligned subspaces while keeping the latent at $g d_h$. The physical split is completed at the BKV stage below: the $d_h^R$ RoPE-only columns are sliced off as the decoupled key $\mathbf{k}^R_t \in \R^{d_h^R}$, and the NoPE remainder is jointly compressed with $V$ into the shared latent $\mathbf{c}^{KV}_t \in \R^{r_{kv}}$. FreqFold arranges coordinates for this split while preserving the frequency-band structure; subsequent rank truncation can introduce approximation error.

\paragraph{$K_{\text{nope}}/V$ norm balancing and joint compression.}
The NoPE part of $K$ and the value tensor $V$ have very different activation norms, so a joint PCA over $[K_{\text{nope}}; V]$ wastes rank on whichever side has larger variance and starves the other. We rescale each side to comparable Frobenius norm and absorb the inverse scale back into $W^{UK}, W^{UV}$ (forward pass unchanged); a single joint PCA then compresses both sides to $r_{kv}$ with rank allocated after balancing the two sides. The retained energy depends on the calibration spectrum.

\paragraph{NLL-weighted activation second moments.}
The GQA route contains two PCA stages: the per-frequency-band FreqFold PCA
and the final joint $[K_{\mathrm{NoPE}};V]$ compression. We apply the same
loss-weighting heuristic to both. Inspired by sensitivity-aware
compression~\citep{lecun1989optimal,frantar2023sparsegpt}, we assign each token
$w_t=\mathrm{NLL}(x_{t+1}\mid x_{\leq t})$ using a frozen teacher forward pass.
The post-norm hidden state is passed through the frozen \texttt{lm\_head};
weights are normalized to unit mean within each batch. For stage-$s$
activations, we accumulate the uncentered second moment
\begin{equation}
\bar\Sigma^{(s)} = \sum_t w_t\, x^{(s)}_{t}\, x^{(s)\,\top}_{t},
\label{eq:nll-second-moment}
\end{equation}
where $s\in\{\mathrm{FreqFold},\mathrm{BKV}\}$.
This is ordinary activation PCA after scaling each token's activation by
$\sqrt{w_t}$. The weighting emphasizes difficult calibration tokens;
it is not a computation of the loss Hessian and has no guaranteed
second-order approximation to the end-to-end loss.

We add Tikhonov damping at
$10^{-2}\operatorname{tr}\bar\Sigma^{(s)}/\dim(x^{(s)})$ and eigendecompose
in fp64. The leading eigenvectors define the frequency-band or joint K/V
projector. We use uncentered statistics to retain the mean activation direction.
The implemented centering ablation differs by at most $0.1$ PPL on LLaMA-3-8B
and Qwen2.5-7B and is $2.86$ PPL worse on DeepSeek-V3.1-Base.
The same weighting applies to the per-group, per-side PCA on the MLA route.
Grouping and weighting can be combined, but their empirical effects need
not be independent or additive (Table~\ref{tab:conversion}).
All stages preserve the query-group assignment established during head merging.

\subsection{From MLA to GQLA}
\label{app:mla-to-gqla}

A pretrained MLA model already has a joint latent, but its head-indexed $W^{UK}, W^{UV}$ do not directly define a small set of shared KV groups. Conversion constructs these groups from the pretrained up-projections. We recover the GQLA factorization in three stages: cluster the $h_q$ query-projected heads into $g$ groups via similarity-based grouping, run per-group side-separated PCA on up-projection activations (optionally NLL-weighted, \S\ref{app:gqa-to-gqla}), and absorb the resulting square reconstruction blocks into $W^{UQ}, W^O$ at the original MLA shapes; the only shape change is $W^{UKV}$ shrinking from $h_q(d_h+d_h^V)$ to $g(d_h+d_h^V)$ rows.

\paragraph{Similarity-based head grouping.}
The MLA latent funnels all $h_q$ query-projected K/V heads through a single subspace, so when we re-expose $g$ groups we must partition the $h_q$ source heads into $g$ target groups of size $h_q/g$. Index-order (``neighbor'') partitioning is the baseline. We instead use calibration statistics to group heads whose activations can be aligned, seeking groups that admit an accurate shared low-rank reconstruction. Streaming a calibration batch through the layer's $W^{UK}, W^{UV}$ chain, we accumulate centered all-pair covariances $\Sigma^{\star} \in \R^{h_q d_h^{\star} \times h_q d_h^{\star}}$ ($\star \in \{K_{\text{nope}}, V\}$), read off per-head-pair blocks $\Sigma^{\star}_{h, h'} \in \R^{d_h^{\star} \times d_h^{\star}}$, and score head similarity by the variance-normalized nuclear norm
\begin{equation}
S[h, h'] = \sum_{\star} w_{\star}\, \frac{\lVert \Sigma^{\star}_{h, h'}\rVert_*}{\sqrt{\operatorname{tr} \Sigma^{\star}_{h, h}\, \operatorname{tr} \Sigma^{\star}_{h', h'}}},
\label{eq:sim-score}
\end{equation}
where the nonnegative side weights $w_{\star}$ balance K and V contributions. The score measures cross-head alignment after variance normalization. The nuclear norm also appears in orthogonal Procrustes alignment, motivating this score as a grouping heuristic. Pairwise affinity is not the same objective as the subsequent multi-head rank-truncated PCA. A balanced greedy clustering (seed each group with the highest-affinity unassigned pair; grow to $h_q/g$ heads by maximum total intra-group affinity) yields a permutation $\pi$ that we apply jointly to the head slices of $W^{UK}, W^{UV}, W^{UQ}, W^{QR}$ and the corresponding input slices of $W^O$, leaving the attention identity intact. This step is specific to MLA$\to$GQLA: the GQA route (\S\ref{app:gqa-to-gqla}) already has $g_{\text{src}}\!=\!g$ KV heads -- there is no $h_q\!\to\!g$ clustering to perform -- so similarity grouping is dormant there. On GLM-4.7-Flash it adds $+1.03$ to $+1.86$ eight-task Avg points (depending on whether NLL-weighted PCA is also enabled; Table~\ref{tab:conversion}).

\paragraph{Per-group, per-side activation PCA.}
The GQA path caches per-group K-NoPE and V independently, so we factor each side $\star \in \{K, V\}$ separately. Partition the rows of $W^{U\star} \in \R^{h_q d_h^{\star} \times r_{kv}}$ into $g$ contiguous query-head groups of size $h_q/g$, yielding blocks $W^{\star}_j \in \R^{D^{\star}_g \times r_{kv}}$ with $D^{\star}_g = (h_q/g)\, d_h^{\star}$. For $N$ calibration tokens, eigendecompose the uncentered per-group activation second moment $\Sigma^{\star}_j = R^{\star}_j \Lambda^{\star}_j R^{\star}_j{}^{\!\top}$, and retain the top $r^{\star} \leq D^{\star}_g$ components:
\begin{equation}
\begin{aligned}
\widetilde{W}^{\star}_j &= U^{\star}_j V^{\star}_j, \\
U^{\star}_j &= R^{\star}_j[:, :r^{\star}], \\
V^{\star}_j &= U^{\star}_j{}^{\!\top} W^{\star}_j,
\end{aligned}
\label{eq:mla2gqla-factor}
\end{equation}
The retained eigenspace minimizes rank-constrained activation reconstruction error under the calibration statistics used to construct $\Sigma^{\star}_j$. Because MLA already supplies the joint latent, this per-group, per-side eigendecomposition is the \emph{only} PCA stage on this route (in contrast to the two stages of \S\ref{app:gqa-to-gqla}). We apply the NLL weighting of \S\ref{app:gqa-to-gqla} here verbatim by substituting $\bar\Sigma^{\star}_j$ from Eq.~\eqref{eq:nll-second-moment} for the uniformly weighted second moment. This yields an NLL-weighted projector for either index-order or similarity grouping.

\paragraph{Canonical rank and weight absorption.}
We set $r^K\!=\!d_h, r^V\!=\!d_h^V$ (one head per side per group, matching GQLA's canonical config). Each $U^{\star}_{j,\ell}$---the $\ell$-th of $m=h_q/g$ stacked $d_h^{\star}\!\times\!d_h^{\star}$ blocks of $U^{\star}_j$---is square and \emph{absorbs} into the per-head $W^{UQ}, W^O$ slices with no shape change. For global query head $i$, let $j(i)=\lceil i/m\rceil$ and
$\ell(i)=i-(j(i)-1)m$ be its group and within-group position:
\begin{align}
&\widetilde{W}^{UKV} \in \R^{g(d_h+d_h^V) \times r_{kv}}\nonumber\\
&\qquad\text{has per-}j\text{ rows } V^K_{j}, V^V_{j}, \nonumber\\
&\widetilde{W}^{UQ}_{i,\,\mathrm{NoPE}} = U^K_{j(i),\,\ell(i)}{}^{\!\top}\, W^{UQ}_{i,\,\mathrm{NoPE}}, \nonumber\\
&\widetilde{W}^{O}_{:,\,i} = W^{O}_{:,\,i}\, U^V_{j(i),\,\ell(i)}. \label{eq:absorb-o}
\end{align}
Substituting Eqs.~\eqref{eq:mla2gqla-factor}--\eqref{eq:absorb-o} into MLA's forward pass recovers GQLA's GQA path up to per-group PCA truncation: $V^{K}_j\mathbf{c}_t^{KV}$ and $V^{V}_j\mathbf{c}_t^{KV}$ supply group $j$'s compressed content keys and values, respectively, and the absorbed $\widetilde{W}^{UQ}, \widetilde{W}^O$ let head $i$ act directly against group $j(i)$'s compressed K-NoPE/V.

\paragraph{Deployment and the dual-path view.}
Absorbing each head's reconstruction blocks into $W^{UQ}$ and $W^O$ leaves
these matrices at their original shapes and yields group-indexed up-projections
$\widetilde{W}^{UKV}$. The converted attention can then use the GQA path of
Eq.~\eqref{eq:mha} or absorb the group projections into the query/output
computations as in Eq.~\eqref{eq:mqa}. These are two execution forms of the
converted model; neither restores information discarded by PCA.
Reusing a head-indexed MLA loader can require repeating each group's
up-projection across its query heads, without changing the represented function.

A per-layer check on GLM-4.7-Flash
($h_q=20$, $d_h=192$, $d_h^V=256$, $r_{kv}=512$, $g=4$), using approximately
$32$ calibration batches, reports an absorbed-versus-unfused PCA BF16 gap of
approximately $10^{-3}$. This is a numerical check of the absorption step,
not a bound on source-to-converted model error. End-to-end conversion quality
is evaluated separately in Table~\ref{tab:conversion}.

\section{H20 and H20-3e: Additional Roofline Comparisons}
\label{app:h20-comparison}

The main text uses H20-3e throughout to match the GPU used in our
kernel and end-to-end serving measurements. Here we retain the H20
analysis and compare the two hardware configurations and cache layouts.
All H20 results in this appendix are analytical predictions;
the measured results reported in the main text use H800 or H20-3e.

\subsection{Hardware parameters}

Table~\ref{tab:hardware-comparison} includes H20 alongside H800 and H20-3e.
Relative to H800, H20 reduces peak BF16 compute by approximately
$6.7\times$ while retaining higher HBM bandwidth, lowering the ridge
from approximately $295$ to $37$ FLOPs/byte.
H20-3e retains H20's $148$ TFLOP/s but raises bandwidth from $4.0$ to
$4.8$ TB/s, lowering the ridge further to approximately $31$ FLOPs/byte.
MQA-absorb has intensity $242$ for $s_q=1$ and $484$ for $s_q=2$.
Both exceed these ridges, so the unsharded model predicts no query-throughput
gain from increasing $s_q$ on either device.

\begin{table}[t]
\centering
\small
\setlength{\tabcolsep}{4pt}
\begin{tabular}{lccc}
\toprule
GPU & BF16 (TFLOP/s) & HBM (TB/s) & Ridge $I^{\star}$ \\
\midrule
H800 & $989$ & $3.35$ & $\approx 295$ \\
H20  & $148$ & $4.0$  & $\approx 37$  \\
H20-3e & $148$ & $4.8$ & $\approx 31$ \\
\bottomrule
\end{tabular}
\caption{BF16 Roofline parameters (dense, no 2:4), from vendor-reported peak specifications rather than our own measurements. H20 and H20-3e have the same peak compute but different HBM bandwidths, giving ridges of approximately $37$ and $31$ FLOPs/byte, respectively.}
\label{tab:hardware-comparison}
\end{table}

\subsection{Shared-RoPE ideal and deployed physical cache layouts}

\begin{figure*}[t]
  \centering
  \begin{subfigure}[t]{0.32\linewidth}
    \centering
    \includegraphics[width=\linewidth]{imgs/roofline_h800.png}
  \end{subfigure}\hfill
  \begin{subfigure}[t]{0.32\linewidth}
    \centering
    \includegraphics[width=\linewidth]{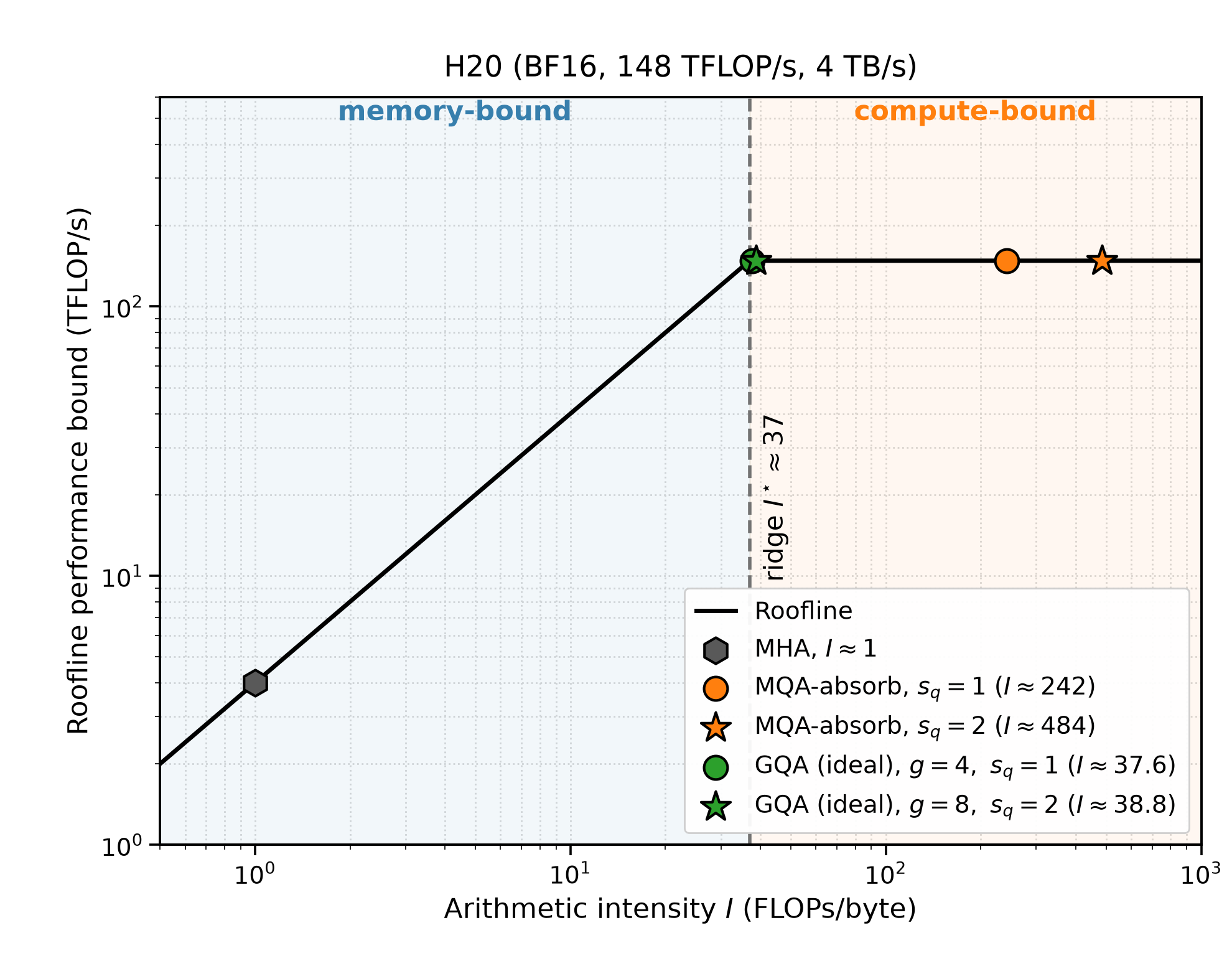}
  \end{subfigure}\hfill
  \begin{subfigure}[t]{0.32\linewidth}
    \centering
    \includegraphics[width=\linewidth]{imgs/roofline_h20_3e.png}
  \end{subfigure}
  \caption{\small
  BF16 decoding Roofline on H800 (left), H20 (middle), and H20-3e (right).
  Solid: $\min(I\!\cdot\!\mathrm{BW}, \mathrm{peak})$;
  dashed: ridge $I^{\star}$.
  On H800, MLA and GQLA share MQA-absorb:
  $s_q\!=\!1$ lands just below, $s_q\!=\!2$ overshoots.
  The H20 GQLA points use the shared-RoPE \emph{ideal} cache accounting,
  whereas the H20-3e points use the deployed dense-GQA \emph{physical} cache accounting.
  Across both H20 panels, MQA-absorb is compute-bound,
  while GQLA's GQA path at $(g, s_q)\!\in\!\{(8,2), (4,1)\}$
  is predicted to sit near the ridge.}
  \label{fig:roofline-comparison}
\end{figure*}

Using the shared-RoPE ideal cache of Eq.~\eqref{eq:gqa-bytes-tok},
the GQA arithmetic intensity is
\begin{equation}
I_{\mathrm{GQA,ideal}} = \frac{h_q s_q (2 d_h + d_h^R)}{2 g d_h + d_h^R}.
\label{eq:gqa-ideal-ai}
\end{equation}
Equation~\eqref{eq:gqa-ideal-ai} uses the shared-RoPE ideal cache accounting.
For $(g,s_q)\!\in\!\{(8,2),(4,1)\}$, it gives
$I_{\mathrm{GQA}}\!\approx\!38.8$ and $37.6$, respectively, placing both near
the theoretical H20 ridge of approximately $37$ FLOPs/byte.
Under the deployed physical layout, the denominator becomes
$g(2d_h+d_h^R)$, so
$I_{\mathrm{GQA,physical}}=h_qs_q/g=32$ for both configurations.
This physical-layout operating point lies similarly close to the
theoretical H20-3e ridge of approximately $31$ FLOPs/byte.
Accordingly, the middle and right panels of Figure~\ref{fig:roofline-comparison}
show the same GQLA designs under ideal and physical cache accounting,
respectively.

The ideal layout uses $4224$ bytes/token/layer at $g=8$ and $2176$
at $g=4$, while the deployed physical layout uses $5120$ and $2560$,
respectively. The $(g,s_q)=(4,1)$ configuration therefore offers a
smaller-cache alternative without MTP, supporting KV sharding without replication
up to TP$=4$; $(8,2)$ supports such sharding up to TP$=8$.

Figure~\ref{fig:roofline-comparison} retains the three-panel comparison.
Its H20 and H20-3e panels vary both hardware bandwidth and cache layout;
the difference between the plotted GQA intensities comes from the layout,
not from the GPU. At a fixed layout, intensity is identical on both devices:
$38.8$ and $37.6$ for the respective ideal-layout configurations and $32$ for both
physical-layout configurations. In the physical layout, these points
are slightly below H20's ridge and slightly above H20-3e's ridge.
The main text consistently uses the physical layout for its operating-point
table and H20-3e predictions.

\section{Roofline Derivations and Full Operating Points}
\label{app:roofline-details}

\subsection{The Roofline model and the H800/H20-3e ridges}
\label{subsec:roofline-model}

The Roofline model~\citep{williams2009roofline} bounds kernel throughput by $\min(I \cdot \mathrm{BW},\, \mathrm{FLOPs}_{\max})$, where $I$ is arithmetic intensity (FLOPs per byte of off-chip traffic). The \emph{ridge} $I^{\star} = \mathrm{FLOPs}_{\max}/\mathrm{BW}$ separates memory- and compute-bound regimes; efficient decoding pushes $I$ toward $I^{\star}$.

MHA decoding has $I\!\approx\!1$~\citep{zadourihardware}, nearly three orders of magnitude below the H800 ridge ($I^{\star}\!\approx\!295$); this gap motivates changes to both attention representations and kernel implementations~\citep{dao2022flashattention,pope2023efficiently}. The export-restricted H20-3e (Table~\ref{tab:hardware}) offers higher HBM bandwidth than H800 but cuts compute $\sim\!7\times$, lowering the ridge to $\sim\!31$. FLOPs have historically outpaced bandwidth~\citep{gholami2024ai}; H800$\to$H20-3e reverses this, so an H800-tuned intensity becomes compute-bound on H20-3e and leaves HBM idle.

\subsection{GQLA on the Roofline}
\label{app:gqla-roofline-details}

We compare the arithmetic intensities of GQLA's two paths with the hardware ridges. These calculations identify favorable attention operating points rather than guarantee achieved throughput. The design space is two paths $\times$ the per-step query count $s_q$ ($s_q\!=\!1$ for ordinary decoding, $s_q\!\geq\!2$ for MTP/speculative). We default to the DeepSeek-V2/V3 canonical $(h_q, g, d_h, d_h^R, r_{kv})\!=\!(128, 8, 128, 64, 512)$. With other dimensions fixed, halving $h_q$ halves both intensities; the preferred operating point must therefore be reconsidered for each model configuration. The DeepSeek-V3.1-Base $g=8$ checkpoint evaluated in
\S\ref{sec:exper} instantiates this canonical configuration,
so the analytical operating points use the same attention dimensions
as our kernel and end-to-end measurements.

\subsubsection{MQA-absorb path: compact latent cache}
\label{subsubsec:mqa-absorb}

MQA-absorb caches only the latent $\mathbf{c}_s^{KV}\!\in\!\R^{r_{kv}}$ and the head-shared RoPE key $\mathbf{k}_s^R\!\in\!\R^{d_h^R}$:
\begin{equation}
\begin{aligned}
N_{\mathrm{MQA}}^{\mathrm{tok}} &= r_{kv} + d_h^R, \\
B_{\mathrm{MQA}}^{\mathrm{tok}} &= 2(r_{kv} + d_h^R)
\end{aligned}
\label{eq:mqa-bytes-tok}
\end{equation}
($1152$ bytes/token at the canonical config). Each step reads all $L$ tokens once and reuses them across the $s_q$ queries (FlashAttention-style), so $B_{\mathrm{MQA}}\!=\!2 L (r_{kv} + d_h^R)$ is independent of $s_q$. After absorption (Eq.~\eqref{eq:mqa}), each (head, query, cache-position) triplet costs $2(2 r_{kv} + d_h^R)$ FLOPs:
\begin{align}
F_{\mathrm{MQA}} &= 2 L\, h_q s_q\, (2 r_{kv} + d_h^R), \label{eq:mqa-flops}\\
I_{\mathrm{MQA}} &= \frac{h_q s_q (2 r_{kv} + d_h^R)}{r_{kv} + d_h^R}. \label{eq:mqa-ai}
\end{align}
$I_{\mathrm{MQA}}$ scales linearly with $s_q$~\citep{deepseekv3_inference}: $s_q\!=\!1$ gives $I\!\approx\!242$ (just below the H800 ridge), $s_q\!=\!2$ gives $484$ (overshoots). DeepSeek-V3 defaults to MTP ($s_q\!=\!2$), so H800 per-step time grows from $2.82$ to $4.61\,\mu\text{s}$ and the ideal query-throughput gain falls from $2\times$ to $\sim\!1.22\times$.

\subsubsection{GQA path: per-group expanded cache}
\label{subsubsec:gqa-path}

The GQA path algorithmically caches per-group expanded $K_C, V$
($g d_h$ each) plus one shared RoPE key. Under this shared-RoPE ideal,
\begin{equation}
\begin{aligned}
N_{\mathrm{GQA,ideal}}^{\mathrm{tok}} &= 2 g d_h + d_h^R, \\
B_{\mathrm{GQA,ideal}}^{\mathrm{tok}} &= 2(2 g d_h + d_h^R).
\end{aligned}
\label{eq:gqa-bytes-tok}
\end{equation}
When deployment reuses an existing dense GQA kernel,
its cache ABI physically repeats the $d_h^R$ RoPE slice across KV groups:
\begin{equation}
\begin{aligned}
N_{\mathrm{GQA,physical}}^{\mathrm{tok}} &= g(2d_h+d_h^R), \\
B_{\mathrm{GQA,physical}}^{\mathrm{tok}} &= 2g(2d_h+d_h^R).
\end{aligned}
\label{eq:gqa-physical-bytes-tok}
\end{equation}
At BF16 and $g=8$, the ideal cache uses $4224$ bytes/token/layer; the physical
cache uses $5120$.
The ideal cache mirrors LLaMA-3 GQA's $2 g d_h\!=\!2048$
plus $d_h^R$ extra elements, but $K, V$ are constrained to the
rank-$r_{kv}$ subspace of GQLA's up-projections,
so expressivity differs from a free same-$d_h$ GQA.
Per-triplet FLOPs are $2(2 d_h + d_h^R)$:
\begin{align}
F_{\mathrm{GQA}} &= 2 L\, h_q s_q\, (2 d_h + d_h^R), \label{eq:gqa-flops}\\
I_{\mathrm{GQA,physical}} &= \frac{h_q s_q (2 d_h + d_h^R)}{g(2 d_h + d_h^R)} = \frac{h_q s_q}{g}. \label{eq:gqa-ai}
\end{align}
Equation~\eqref{eq:gqa-ai} uses the deployed physical cache layout,
matching the kernel and serving measurements.
For $(g,s_q)\!\in\!\{(8,2),(4,1)\}$, it gives
$I_{\mathrm{GQA,physical}}=32$, close to the theoretical H20-3e ridge
of approximately $31$ FLOPs/byte (Figure~\ref{fig:roofline}, right).
Additional hardware and cache-layout comparisons appear in
Appendix~\ref{app:h20-comparison}.

\subsubsection{Operating points across hardware}
\label{subsubsec:operating-points}

\begin{table*}[t]
\centering
\small
\setlength{\tabcolsep}{4pt}
\textbf{Analytical bounds at TP$=1$}
\par\smallskip
\resizebox{\textwidth}{!}{%
\begin{tabular}{llcccccccc}
\toprule
GPU & Path & $n_{kv}$ & $s_q$ & \shortstack{Cache\\(B/token)} & $I$ & \shortstack{Memory\\($\mu$s)} & \shortstack{Compute\\($\mu$s)} & \shortstack{Step\\($\mu$s)} & \shortstack{Theory\\(query/s)} \\
\midrule
\multirow{6}{*}{H800} & \textbf{MQA-absorb} & \textbf{1} & \textbf{1} & $1152$ & $\mathbf{242}$ & $2.82$ & $2.31$ & $\mathbf{2.82}$ & $\mathbf{355.0\rm K}$ \\
 & MQA-absorb & 1 & 2 & $1152$ & $484$ & $2.82$ & $4.61$ & $4.61$ & $433.4\rm K$ \\
 & GQA & 8 & 1 & $5120$ & $16$ & $12.52$ & $0.68$ & $12.52$ & $79.9\rm K$ \\
 & GQA & 8 & 2 & $5120$ & $32$ & $12.52$ & $1.36$ & $12.52$ & $159.7\rm K$ \\
 & GQA & 4 & 1 & $2560$ & $32$ & $6.26$ & $0.68$ & $6.26$ & $159.7\rm K$ \\
 & GQA & 4 & 2 & $2560$ & $64$ & $6.26$ & $1.36$ & $6.26$ & $319.5\rm K$ \\
\midrule
\multirow{6}{*}{H20-3e} & MQA-absorb & 1 & 1 & $1152$ & $242$ & $1.97$ & $15.42$ & $15.42$ & $64.9\rm K$ \\
 & MQA-absorb & 1 & 2 & $1152$ & $484$ & $1.97$ & $30.83$ & $30.83$ & $64.9\rm K$ \\
 & GQA & 8 & 1 & $5120$ & $16$ & $8.74$ & $4.53$ & $8.74$ & $114.4\rm K$ \\
 & \textbf{GQA} & \textbf{8} & \textbf{2} & $5120$ & $\mathbf{32}$ & $8.74$ & $9.07$ & $\mathbf{9.07}$ & $\mathbf{220.5\rm K}$ \\
 & \textbf{GQA} & \textbf{4} & \textbf{1} & $2560$ & $\mathbf{32}$ & $4.37$ & $4.53$ & $\mathbf{4.53}$ & $\mathbf{220.5\rm K}$ \\
 & GQA & 4 & 2 & $2560$ & $64$ & $4.37$ & $9.07$ & $9.07$ & $220.5\rm K$ \\
\bottomrule
\end{tabular}}
\par\medskip
\textbf{Measured kernel throughput (query positions/s)}
\par\smallskip
\begin{tabular}{llcccccc}
\toprule
GPU & Path & $n_{kv}$ & $s_q$ & TP$=1$ & TP$=2$ & TP$=4$ & TP$=8$ \\
\midrule
\multirow{6}{*}{H800} & \textbf{MQA-absorb} & \textbf{1} & \textbf{1} & 261.3K & 306.0K & 310.2K & 312.7K \\
 & MQA-absorb & 1 & 2 & 272.3K & 517.4K & 610.3K & 619.7K \\
 & GQA & 8 & 1 & 74.7K & 148.1K & 291.3K & 558.2K \\
 & GQA & 8 & 2 & 148.0K & 294.4K & 580.0K & 1.106M \\
 & GQA & 4 & 1 & 147.5K & 290.1K & 553.7K & 558.5K \\
 & GQA & 4 & 2 & 293.6K & 567.0K & 1.090M & 1.105M \\
\midrule
\multirow{6}{*}{H20-3e} & MQA-absorb & 1 & 1 & 47.4K & 94.3K & 95.0K & 95.4K \\
 & MQA-absorb & 1 & 2 & 54.0K & 94.5K & 188.6K & 190.1K \\
 & GQA & 8 & 1 & 100.6K & 194.0K & 376.0K & 742.0K \\
 & \textbf{GQA} & \textbf{8} & \textbf{2} & 145.2K & 275.1K & 538.7K & 1.045M \\
 & \textbf{GQA} & \textbf{4} & \textbf{1} & 136.3K & 266.5K & 517.6K & 737.7K \\
 & GQA & 4 & 2 & 117.2K & 203.9K & 408.3K & 1.053M \\
\bottomrule
\end{tabular}
\caption{Analytical and measured per-step operating points
($L\!=\!8192$, BF16, canonical config).
Measured MQA-absorb and GQA rows use the existing FA3-MLA and HPC-Ops
kernels, respectively.
Here $n_{kv}$ counts unsharded cache heads: one latent head for MQA-absorb,
and $g$ expanded groups for GQA. It does not redefine the model's $g$.
The cache, intensity, and Roofline columns use TP$=1$ and the physical layout;
the measured panel uses $B=128$ and simulates rank-local TP shapes on one GPU.
All throughputs count query positions, including both queries at $s_q=2$.
Time $=\max(F/\mathrm{FLOPs}_{\max}, B/\mathrm{BW})$ and theoretical
throughput $=s_q/\text{step}$.
At $(h_q,g)\!=\!(128,8)$, H800 MQA-absorb gives $355.0$K tok/s at $s_q=1$,
while H20-3e GQA gives $220.5$K tok/s at $s_q=2$;
$(g,s_q)\!=\!(4,1)$ is an equally
ridge-optimal H20-3e alternative.}
\label{tab:roofline}
\end{table*}

Table~\ref{tab:roofline} tabulates step time across
hardware $\times$ path $\times s_q$.
Three takeaways: (1) on H800, MQA-absorb at $s_q\!=\!1$ is fastest
($2.82\,\mu\text{s}/$step); MTP makes it compute-bound and shrinks the
speedup to $1.22\times$ in the TP$=1$ analytical model;
(2) at TP$=1$, MQA-absorb is compute-bound on H20-3e for both $s_q=1$ and $2$,
so the model predicts no query-throughput gain from increasing $s_q$; (3) GQLA's GQA path at
$(g, s_q)\!\in\!\{(8,2), (4,1)\}$ is predicted to reach the H20-3e ridge
at $221$K tok/s---$3.4\times$ over MQA-absorb on the same device in the analytical
model. A fixed GQLA checkpoint supports both paths without retraining or
custom kernels; changing $g$ requires a different grouping and weights.
The H20-3e path-selection prediction has been validated by on-device
measurements, as detailed in \S\ref{sec:exper}.

\subsubsection{Choosing \texorpdfstring{$(g, s_q)$}{(g, sq)}}
\label{subsubsec:gsq-choice}

The two near-ridge configurations trade expanded-cache size, group-sharding
capacity, and queries per step. The analysis favors $(g,s_q)\!=\!(8,2)$: a wider
group-expanded representation ($g d_h=1024$ versus joint latent rank
$r_{kv}=512$) with KV sharding without replication up to TP$=8$.
The lighter H20-3e alternative $(g,s_q)\!=\!(4,1)$ halves the physical cache
to $2560$ bytes/token and needs no MTP head, but has square
$W^{UK}\!\in\!\R^{512\times512}$ and supports such sharding only up to TP$=4$.
Both group counts remain deployable on H800 at $2.82\,\mu\text{s}$/step
with MQA-absorb and $s_q=1$, since $I_{\mathrm{MQA}}$ is $g$-independent.
The $g=4,s_q=2$ row is compute-bound on H20-3e in this model.
These are analytical comparisons; the deployed DeepSeek checkpoint uses $g=8$.
We do not evaluate its conversion quality at $g=4$ or alternative latent ranks.

\section{Experimental Details and Full Conversion Results}
\label{app:exper-details}

We organize the empirical evaluation into four parts.
First, we compare MLA and GQLA under a shared pretraining recipe
at multiple model sizes.
Second, we evaluate TransGQLA on four pretrained backbones spanning both
conversion routes of \S\ref{sec:transgqla}: LLaMA-3-8B~\citep{grattafiori2024llama}
and Qwen2.5-7B~\citep{qwen2_5_2025} via GQA$\to$GQLA, and
GLM-4.7-Flash~\citep{glm47} and DeepSeek-V3.1-Base~\citep{liu2024deepseekv3}
via MLA$\to$GQLA.
For all four, we measure the quality retained immediately after the
zero-training-token architectural conversion.
For LLaMA-3-8B, Qwen2.5-7B, and GLM-4.7-Flash, we additionally evaluate
continued pretraining and supervised fine-tuning to determine how fully
the remaining conversion gap can be recovered.
Third, we validate the predicted H20-3e path selection using decode-kernel
measurements of the existing FA3-MLA and HPC-Ops GQA implementations
across rank-local TP shapes.
Fourth, we compare both paths in TP$=8$ full-model serving on
DeepSeek-V3.1-Base across 8K--64K inputs, under matched resident batches
and independently capacity-tuned settings.

\subsection{Scaling with Pretraining Compute}
\label{app:scaling-details}

\paragraph{Setup.}
We pretrain MoE transformers on a shared multilingual mixture of text and code,
using the same tokenizer, sequence length $8192$, and full attention.
The first layer uses a dense feed-forward network;
subsequent layers use $256$ routed experts with top-$8$ routing and one shared expert.
The KV latent rank is $512$, the non-RoPE QK and value head dimensions are $128$,
and the RoPE dimension is $64$.
We train in BF16 with distributed Muon, weight decay $0.1$,
gradient clipping at $1.0$, and seed $1234$.
The learning rate follows a warmup--stable--decay schedule,
with arccosine decay over the final $30\%$ of training and a minimum of $10^{-6}$.

\paragraph{Scope of the scaling evidence.}
The labels 1:4 and 1:8 specify query heads per KV group ($m$), with
$g=h_q/m$ at each size. The plotted endpoints average the final 100 training
iterations; they are neither held-out losses nor independent repeated runs.
Equation~\eqref{eq:scaling-law} fits three parameters independently for each
method: six endpoints leave three residual degrees of freedom for MLA and
1:8, while four leave only one for 1:4. The fits have no reported uncertainty
intervals, and the dashed extensions to $10^{23}$ FLOPs are extrapolations.
The close exponents describe this training ladder; they do not establish equal
asymptotic laws, compute-optimal training, or downstream generalization.
Matched-token loss gaps compare models within a size; the compute curves also
depend on the architecture-specific FLOP estimates used for the horizontal axis.

\subsection{Model Conversion and Recovery}

\paragraph{Setup.}
\emph{LLaMA-3-8B (GQA$\to$GQLA)} ($h_q\!=\!32$, $g\!=\!8$, $d_h\!=\!128$; original cache $2 g d_h\!=\!2048$ BF16 elts/tok). We apply \S\ref{sec:gqa-to-gqla}: a GQA-preserving head merge retains both paths, followed by TransMLA-style RoRoPE, FreqFold, and activation-balanced low-rank compression~\citep{meng2026transmla} into the canonical latent ($r_{kv}\!=\!512$, $d_h^R\!=\!64$; $576$ cached elts/tok, $28.125\%$ of the GQA baseline). The GQA-path cache is $2 g d_h + d_h^R$, comparable to the original. Calibration: $128$ wikitext-2 samples of seqlen $256$.
\emph{Qwen2.5-7B (GQA$\to$GQLA)} ($h_q\!=\!28$, $g_{\text{src}}\!=\!4$, $d_h\!=\!128$; original cache $2 g_{\text{src}} d_h\!=\!1024$ BF16 elts/tok). Same pipeline at $g\!=\!4$, $r_{kv}\!=\!512$, $d_h^R\!=\!64$; cached $576$ elts/tok ($56.25\%$ of the GQA baseline). Same calibration setup.
\emph{GLM-4.7-Flash (MLA$\to$GQLA).} We apply the per-group, side-separated activation PCA of \S\ref{sec:mla-to-gqla} at canonical rank $r^K\!=\!d_h$, $r^V\!=\!d_h^V$, augmented with similarity grouping and NLL-weighted PCA (\S\ref{sec:gqa-to-gqla},~\ref{sec:mla-to-gqla}); the per-group factors absorb back into $W^{UQ}, W^O$ without shape changes, and $W^{UKV}$ shrinks from $h_q(d_h+d_h^V)$ to $g(d_h+d_h^V)$ rows ($h_q\!=\!20, g\!=\!4$). The MQA-absorb path inherits MLA's original latent cache unchanged; the GQA path uses the expanded cache of \S\ref{subsubsec:gqa-path}. Because the two paths are algebraically equivalent (\S\ref{sec:gqla}), we report one accuracy per row. Calibration: $128$ wikitext-2 samples of seqlen $512$.
\emph{DeepSeek-V3.1-Base (MLA$\to$GQLA).} The headline DeepSeek-family backbone~\citep{liu2024deepseekv3} ($h_q\!=\!128$, $g\!=\!8 \Rightarrow 16$ source heads per KV group, $d_h\!=\!128$, $d_h^V\!=\!128$, $r_{kv}\!=\!512$, $d_h^R\!=\!64$, $q_{\mathrm{LoRA}}\!=\!1536$; $61$ backbone layers + $1$ MTP head (the latter auto-skipped by the HF DeepseekV3 loader), MoE with $256$ routed + $1$ shared expert per MoE layer at $\mathrm{moe\_intermediate}\!=\!2048$, FP8 blockwise-quantized checkpoint at $642$ GB on-disk). The recipe is the same per-group activation PCA of \S\ref{sec:mla-to-gqla} at canonical rank, augmented with similarity grouping and NLL-weighted PCA. We use $g\!=\!8$ ($h_q/g\!=\!16$, the canonical grouping analysed in \S\ref{sec:roofline}) for a $16\!\times$ reduction in head-expanded K/V rows, while retaining the source joint latent. Calibration: $128$ wikitext-2 samples of seqlen $512$, same as GLM-4.7. The $671$B / $642$ GB FP8 weights are loaded CPU-resident via transformers' \texttt{FineGrainedFP8HfQuantizer}; for both calibration and PCA-and-absorb we shuttle one decoder layer at a time to a single $80$ GB GPU (the FP8 Triton w8a8 matmul handles dequant on-chip), and the absorbed attention is written back as plain BF16 \texttt{nn.Linear} (the four modified attention projections occupy $\sim\!350$ MB/layer after absorption and are stored in BF16). The MoE FP8 weights are unchanged and symlinked from the source checkpoint at save time; the converted checkpoint adds only the BF16 attention overrides ($\sim\!21$ GB).
For recovery, our single-seed LLaMA-3-8B, Qwen2.5-7B, and
GLM-4.7-Flash runs use, respectively, 30B, 1B, and 12B tokens of continued
pretraining on ProLong, followed by full-parameter assistant-only SFT on
UltraChat-200K at sequence length 4096, global batch 128, and peak learning
rate $10^{-5}$.

\paragraph{Benchmarks.}
We report wikitext-2 perplexity (lower is better) and zero-shot accuracy on an eight-task suite of reasoning and knowledge benchmarks.
For LLaMA-3-8B, Qwen2.5-7B, and GLM-4.7-Flash, we additionally evaluate
checkpoints after continued pretraining and supervised fine-tuning.
The headline comparison (Table~\ref{tab:conversion}) reports the full $8$-task
suite (PIQA, Winogrande, ARC-easy, ARC-challenge, BoolQ, MMLU, HellaSwag,
OpenBookQA), with \textbf{Avg} the unweighted $8$-task mean so that the numbers
are comparable across all variants.
We use the standard lm-evaluation-harness metric per task: \texttt{acc} for
MMLU, \texttt{acc\_norm} for HellaSwag and OpenBookQA, \texttt{acc} for the
remaining tasks.

\paragraph{Cost of zero-training conversion.}
The zero-training-token rows of Table~\ref{tab:conversion} report the pure
architectural transformation before CPT/SFT.
\emph{MLA$\to$GQLA} on GLM-4.7-Flash (full similarity$+$NLL weighting pipeline)
loses $3.55$ Avg.\ pts ($71.77\!\to\!68.22$) while PPL improves from
$11.39$ to $11.10$, retaining $68.22/71.77=95.0\%$ of the MLA teacher's
$8$-task Avg without training.
\emph{GQA$\to$GQLA} on LLaMA-3-8B loses $12.08$ Avg.\ pts
($69.02\!\to\!56.94$) while PPL rises from $6.14$ to $18.57$ under the
aggressive $3.56\times$ cache compression
($2048\!\to\!576$ elements/token).
The near-identical TransMLA and vanilla GQA$\to$GQLA results
(PPL $25.77$ vs.\ $25.87$; Avg $56.03$ vs.\ $56.14$) indicate that the
dominant conversion loss comes from the joint $K,V$ subspace compression
inherited from TransMLA, rather than from the group-indexed up-projections.

\paragraph{Both-routes comparison (Table~\ref{tab:conversion}).}
On LLaMA-3-8B and Qwen2.5-7B, plain GQA$\to$GQLA differs from TransMLA by
at most $0.4\%$ in PPL and $0.11$ percentage points in Avg.
Adding NLL weighting changes PPL from $25.87$ to $18.57$ on LLaMA and
from $8.41$ to $8.25$ on Qwen. The larger gain on LLaMA accompanies a
larger initial compression loss, but these observations do not isolate
which model properties cause the difference.

On GLM-4.7-Flash, NLL weighting alone gives the lowest PPL ($10.41$),
whereas grouping plus weighting gives the highest converted Avg ($68.22$).
Grouping adds $1.86$ Avg points without weighting and $1.03$ with it;
weighting adds $0.86$ points without grouping and $0.03$ with it.
Thus the refinements can be combined, but their measured gains are not
additive. The combined configuration remains $3.55$ points below the
source model's $71.77$. Small differences between configurations are
reported descriptively, without a claim of statistical significance.

\paragraph{DeepSeek-V3.1-Base (Table~\ref{tab:conversion}).}
We apply the recipe to DeepSeek's 671B-parameter FP8 MoE with 61 layers and
128 query heads. The transformer quantizer stores \texttt{float8\_e4m3fn}
weights in approximately $650$ GB of host RAM; layers are shuttled to one GPU,
where the Triton w8a8 kernel dequantizes on-chip.
At $g=8$, each group shares 16 source heads, reducing expanded K/V rows by
$16\times$. The resulting difference between PPL and task accuracy is more
pronounced than on GLM-4.7-Flash.
\textbf{The combined refinements give the lowest PPL}: plain
MLA$\to$GQLA has PPL $8.19$, leaving an
$8.19\!-\!3.21\!=\!4.98$ PPL gap to the teacher.
NLL weighting alone changes PPL to $8.32$, whereas grouping alone lowers it
to $6.21$ and adding NLL weighting further lowers it to $\mathbf{6.05}$
($1.89\!\times$ teacher PPL).
Measured as closure of the teacher gap, grouping alone closes $\sim\!40\%$
and the full pipeline closes $\sim\!43\%$.
Equivalently, relative to the plain-conversion PPL, the full pipeline is a
$26\%$ reduction ($(8.19\!-\!6.05)/8.19$); we quote the teacher-gap and
baseline-PPL denominators separately to avoid conflating them.
\textbf{Task accuracy varies much less than PPL}: all four converted cells
land within ${\sim}0.09$ pp of each other on the $8$-task average
($77.78$--$77.86$) and within $0.07$ pp of the MLA teacher's $77.84$.
The task averages therefore change much less than perplexity after grouping
$128$ source heads into $8$ KV groups. This supports preservation of the
measured multiple-choice scores, but does not establish equivalence in
next-token prediction or in tasks outside this suite.
The recommended deployment point is \textbf{$+$ grouping $+$ NLL-weighted PCA}:
it has the lowest PPL of the four ($\mathbf{6.05}$), while its $77.79$
eight-task average remains within $0.05$ pp of the teacher.
Each conversion configuration takes approximately $3$ h on one $80$ GB GPU,
including PPL evaluation and saving. Teacher activations are cached across
configurations. Compression takes approximately $2.5$ min with index-order
grouping and $2.3$ h with similarity grouping; the saved checkpoint adds only $21$ GB of BF16 attention
overrides and symlinks the $642$ GB MoE FP8 shards.
The vLLM task evaluation (TP$=8$) runs the MQA-absorb decode kernel
with the same tensor layout as vanilla DeepSeek-V3 (the GQLA
\texttt{kv\_b\_proj} is expanded back to MLA layout by
\texttt{repeat\_interleave} at weight-load time).

\paragraph{Post-conversion recovery.}
The converted initialization produced by TransGQLA enables recovery toward
source-model performance with the reported CPT and SFT budgets.
As shown in Table~\ref{tab:conversion}, on LLaMA-3-8B, 30B CPT followed by
$\sim$47M SFT reduces PPL from $18.57$ to $7.06$ and raises the $8$-task Avg
from $56.94$ to $69.02$, matching the source model's $69.02$.
On Qwen2.5-7B, 1B CPT followed by $\sim$47M SFT reduces PPL from $8.25$ to
$7.29$ and raises Avg from $67.17$ to $70.61$, compared with $70.39$ for the
source model.
On GLM-4.7-Flash, 12B CPT followed by $\sim$47M SFT reduces PPL from $11.10$
to $6.73$ and raises Avg from $68.22$ to $71.64$, compared with $71.77$ for
the MLA teacher.
Across the three recovery runs, the final $8$-task averages are within
$0.22$ percentage points of their source models, but individual task gaps remain: LLaMA's MMLU score is
$57.61$ versus the source's $62.07$.

\paragraph{Numerical absorption check.}
The GLM-4.7-Flash per-layer check reports an absorbed-versus-unfused PCA
BF16 gap of approximately $10^{-3}$ with approximately $32$ calibration
batches (Appendix~\ref{app:mla-to-gqla}). This checks the numerical effect
of reparameterizing the PCA factors. The task results in Table~\ref{tab:conversion}
measure overall conversion quality and do not isolate accumulation of
rounding error across the $47$ layers.

\subsection{Kernel-Level Validation}
\label{app:kernel-details}

Table~\ref{tab:roofline} compares the decode-kernel throughput of GQLA's
two execution paths across rank-local shapes corresponding to
TP$=1,2,4,8$. The MQA-absorb path uses the existing FA3-MLA paged kernel,
while the GQA path uses the existing HPC-Ops adaptive split-K kernel.
The benchmark uses $B\!=\!128$, $L\!=\!8192$, BF16 queries and KV caches,
and page size $64$ on one GPU of the reported hardware type; each invocation is measured with
cold L2 after kernel initialization and correctness checks. The TP columns
simulate the corresponding per-rank shapes on one GPU and therefore include
neither NCCL communication nor multiplication of throughput by TP. These
measurements isolate attention-kernel behavior and do not directly represent
end-to-end serving throughput, which additionally depends on the rest of the
model, tensor-parallel communication, KV-cache management, and scheduling;
\S\ref{subsec:end-to-end-serving} reports the corresponding full-model serving results.

At the TP$=8$ rank-local shape (16 query heads and one KV head per rank),
HPC-Ops GQA reaches $742.0$K tok/s for $s_q\!=\!1$ and $1.045$M tok/s
for $s_q\!=\!2$, compared with $95.4$K and $190.1$K tok/s for FA3-MLA,
corresponding to $7.78\times$ and $5.50\times$ speedups, respectively.
On H800, MQA-absorb is faster at TP$=1$, but the ranking reverses at
TP$=8$: for $s_q=1$, HPC-Ops GQA reaches $558.2$K query positions/s
versus FA3-MLA's $312.7$K; for $s_q=2$, the rates are $1.106$M and $619.7$K,
a $1.79\times$ advantage at both query counts.
TP changes the two paths' rank-local operating points differently.
For MQA-absorb, query-side FLOPs decrease with TP while the latent KV cache
remains replicated. TP$=8$ reduces intensity from $242$ to $30.2$ FLOPs/byte
for $s_q=1$, and from $484$ to $60.4$ for $s_q=2$.
For physical GQA, FLOPs and KV-cache traffic are sharded together, preserving
intensity at $16$ FLOPs/byte for $s_q=1$ and $32$ for $s_q=2$, while
reducing the cache from $5120$ to $640$ B/token per rank.
Against the theoretical H20-3e ridge of approximately $31$ FLOPs/byte,
the TP$=8$ GQA path is memory-bound at $s_q\!=\!1$ and remains close to the
ridge at $s_q\!=\!2$, whereas MQA-absorb remains compute-bound at
$s_q\!=\!2$.
The resulting theoretical GQA-over-MQA-absorb speedups are $1.80\times$ and
$3.40\times$ at TP$=8$, compared with $1.76\times$ and $3.40\times$
at TP$=1$.
The larger kernel-level gains therefore combine this structural advantage
with the different efficiencies of the reused kernels at rank-local shapes.

\subsection{End-to-End Serving}
\label{app:serving-details}

\paragraph{Setup.}
Using \texttt{vllm bench serve}, we compare the MQA-absorb and GQA
execution paths of the converted DeepSeek-V3.1-Base $g=8$ checkpoint
in full-model online serving. Experiments run on a single node with
eight 141\,GB H20-3e GPUs using TP$=8$ and PP$=1$. The MQA-absorb
path uses FA3-MLA, while the GQA path uses HPC-Ops; for each
matched-resident-batch comparison, all other model and serving settings
are held fixed. We evaluate 8K, 16K, 32K, and 64K inputs with a fixed
output length of 512 tokens, BF16 compute and KV cache, PIECEWISE CUDA Graph and
chunked prefill enabled, GPU-memory utilization of $0.97$, no CPU
offload, and no MTP. The primary comparison uses matched resident
batches of $B=64,40,20,10$, respectively, and we separately compare
independently capacity-tuned near-capacity configurations at 16K.
Each run submits $N=2B$ requests at client concurrency $C=2B$,
forming a two-wave online burst, with seed $42$ and $16$ client warmups
excluded from measurement. Each setting uses five runs with a fresh server
per run, and all reported metrics are averaged over these five runs.
We report aggregate output throughput, mean TPOT, TTFT, and end-to-end
latency, together with the calculated per-GPU live KV-cache footprint.

\paragraph{Matched workloads.}
Table~\ref{tab:serving} reports the full-model serving results under
matched resident batches. Across 8K--64K inputs, the GQA path improves aggregate
output throughput by $22.42$--$34.97\%$, with the largest gain of
$34.97\%$ at 16K. These throughput improvements are accompanied by
$21.18$--$30.29\%$ lower mean TPOT, $11.66$--$18.13\%$ lower mean TTFT,
and $16.90$--$24.31\%$ lower mean end-to-end latency. Although the
end-to-end gains are smaller than the isolated kernel speedups in
\S\ref{app:kernel-details}, they persist after full-model prefill
and decode, MoE computation, TP communication, KV-cache management,
and scheduling are included. Under TP$=8$, the sharded GQA cache also
reduces the live KV-cache footprint on each GPU by $44.4\%$ at every
matched point; at 16K, for example, it decreases from $44.23$\,GiB for
MQA-absorb to $24.57$\,GiB for GQA. This additional memory headroom motivates
the independently capacity-tuned comparison below.

\paragraph{KV capacity and near-capacity throughput.}
Under TP$=8$, MQA-absorb retains its replicated $1152$-B latent cache on every
rank, whereas GQA stores the $640$-B rank-local shard of the deployed
physical GQA cache. At the same KV-memory budget, this gives GQA
$1.8\times$ the token capacity. To evaluate the resulting
memory headroom, we independently tune the two paths near their
respective KV-capacity limits at 16K. MQA-absorb uses $B\!=\!40$ and reaches
$143.31$ output tok/s, while GQA increases the resident batch to
$B\!=\!72$ and reaches $215.05$ output tok/s; both configurations have
a calculated live KV footprint of $44.23$\,GiB per GPU. Increasing GQA
from $B\!=\!40$ to $B\!=\!72$ adds $11.18\%$ over its matched-B
throughput. Combined with GQA's $34.97\%$ matched-batch advantage over
MQA-absorb, the independently capacity-tuned comparison reaches a $50.07\%$
full-model output-throughput gain.

\end{document}